\documentclass{article} % For LaTeX2e
\usepackage{colm2024_conference}

\usepackage{microtype}
\usepackage{hyperref}
\usepackage{url}
\usepackage{booktabs}

\title{What makes a good metric?\\ Evaluating automatic metrics for text-to-image consistency}

\author{Candace Ross, Melissa Hall, Adriana Romero Soriano, Adina Williams\\ %\thanks{ Use footnote for providing further information about author (webpage, alternative address)---\emph{not} for acknowledging funding agencies.  Funding acknowledgements go at the end of the paper.} \\
Meta AI (FAIR)\\
\texttt{\{ccross,melissahall,adrianars,adinawilliams\}@meta.com} \\
}

\usepackage{graphicx}
\graphicspath{{images/}{images/ablations/}{images/ablations/ablations-bar-plots}{images/heatmaps/}}

\usepackage{enumitem}
\usepackage{hyperref}
\usepackage{graphicx}
\usepackage{caption}
\usepackage{subcaption}
\usepackage{booktabs}
\usepackage{xcolor}
\usepackage{multirow}
\usepackage{stfloats} % for positioning of figure* or table* on the same page
\usepackage{csvsimple}
\usepackage{datatool}
\usepackage{pgfplotstable}
\pgfplotsset{compat=1.18}
\usepackage{filecontents}
\usepackage{tikz}
\usetikzlibrary{positioning}
\usetikzlibrary{fit}
\usepackage{arydshln}
\usepackage{amssymb}
\usepackage{pifont}

\usepackage{fontawesome5}
\usepackage[normalem]{ulem}

\newcommand{\cmark}{\ding{51}}
\newcommand{\xmark}{\ding{55}}

\definecolor{Red}{rgb}{1,0,0}
\definecolor{Green}{rgb}{0,0.69,0}
\definecolor{Blue}{rgb}{0,0,1}
\definecolor{LightBlue}{rgb}{0,0.5,1}
\definecolor{veryLightBlue}{rgb}{0.85,0.98,1}
\definecolor{veryLightGreen}{rgb}{0.6,1,0.6}
\definecolor{Skin}{rgb}{1,0.71,0.69}
\definecolor{Grey}{rgb}{0.5,0.5,0.5}
\definecolor{LightGrey}{rgb}{0.6,0.6,0.6}
\definecolor{VeryLightGrey}{RGB}{219, 219, 219}
\definecolor{Black}{rgb}{0,0,0}
\definecolor{White}{rgb}{1,1,1}
\definecolor{brickred}{rgb}{0.8, 0.25, 0.33}
\definecolor{burntOrange}{RGB}{255,122,20}
\definecolor{navy}{RGB}{80, 74, 255}
\definecolor{teal}{RGB}{0, 123, 159}
\definecolor{aquamarine}{RGB}{51, 153, 255}
\definecolor{saffron}{RGB}{227, 170, 0}
\definecolor{purplePink}{RGB}{160, 89, 107}
\definecolor{xanadu}{RGB}{126, 145, 129}
\newcommand{\red}{\color{Red}}
\newcommand{\green}{\color{Green}}

\newcommand{\burntOrange}{\color{burntOrange}}

\newcommand{\orange}{\color{orange}}

\newcommand{\eg}{\emph{e.g.}\ }
\newcommand{\ie}{\emph{i.e.}\ }
\newcommand{\TODO}[1]{ \textbf{\red TODO {#1} }}

\newcommand{\glide}{GLIDE\ }
\newcommand{\glideNoSpace}{GLIDE}

\newcommand{\ldmOne}{LDM v1\ }
\newcommand{\ldmTwo}{LDM v2\ }
\newcommand{\ldmOneNoSpace}{LDM v1}
\newcommand{\ldmTwoNoSpace}{LDM v2}

\newcommand{\dmwithclipNoSpace}{DM w/ CLIP latents}
\newcommand{\dmwithclipshort}{DM w/ CLIP\ }

\newcommand{\faithfulness}{text-image consistency\ }

\newcommand{\metric}{text-image consistency metric\ }
\newcommand{\metrics}{text-image consistency metrics\ }
\newcommand{\Metric}{Text-image consistency metric\ }

\newcommand{\metricNoSpace}{text-image consistency metric}
\newcommand{\metricsNoSpace}{text-image consistency metrics}

\definecolor{barplotBlue}{RGB}{43, 92, 255}
\definecolor{barplotOrange}{RGB}{252, 147, 30}
\definecolor{barplotGreen}{RGB}{64, 151, 83}
\definecolor{barplotRed}{RGB}{211, 16, 7}
\definecolor{barplotPurple}{RGB}{115, 64, 151}
\newcommand{\barplotBlue}{\color{barplotBlue}}
\newcommand{\barplotOrange}{\color{barplotOrange}}
\newcommand{\barplotGreen}{\color{barplotGreen}}
\newcommand{\barplotRed}{\color{barplotRed}}
\newcommand{\barplotPurple}{\color{barplotPurple}}

\colmfinalcopy % Uncomment for camera-ready version, but NOT for submission.
\begin{document}

\maketitle

\begin{abstract}
Language models are increasingly being incorporated as components in larger AI systems for various purposes, from prompt optimization to automatic evaluation. In this work, we analyze the construct validity of four recent, commonly used methods for measuring text-to-image consistency---CLIPScore, TIFA, VPEval, and DSG---which rely on language models and/or VQA models as components. We define construct validity for \metrics as a set of desiderata that \metrics should have, and find that no tested metric satisfies all of them.
We find that metrics lack sufficient sensitivity to language and visual properties.
Next, we find that TIFA, VPEval and DSG contribute novel information above and beyond CLIPScore, but also that they correlate highly with each other.
We also ablate different aspects of the \metrics and find that not all model components are strictly necessary, also a symptom of insufficient sensitivity to visual information.
Finally, we show that all three VQA-based metrics likely rely on familiar text shortcuts (such as \textit{yes}-bias in QA) that call their aptitude as quantitative evaluations of model performance into question.

\end{abstract}

\newcommand{\tableCell}[1]{\rotatebox{0}{ #1}}
\newcommand{\leftmostCell}[1]{\scalebox{1}{\parbox{19em}{#1}}}

\section{Introduction}

Text-to-image (T2I) models are becoming increasingly prevalent, leading to a surge in high-quality generated images \citep{nichol2021glide,ramesh2022hierarchical,yu2023scaling}. T2I models take text prompts like ``\textit{the purple dog is laying across a flower bed}'' as input, and generate images that, ideally, will not only be aesthetically pleasing, but also consistent with the text. For example, if the image generated contains a dog, but the dog is not purple nor laying in a flower bed, the generation would be incomplete in an important way. %To determine whether generated images are consistent with their prompts,
Several evaluation frameworks have recently been devised to automatically evaluate this relationship, \ie the \textit{consistency} between the text prompt and the generated image. 

One metric for evaluating the text-image consistency is CLIPScore \citep{hessel2021clipscore}, which uses a CLIP model \citep{radford2021learning} to compute a similarity score between the text caption and the image. Because CLIP does struggle with aspects of visiolinguistic reasoning such as compositionality \citep{thrush2022winoground,yuksekgonul2022and,yu-etal-2023-crepe}, other recent automatic metrics take a more fine-grained approach \citep{hu2023tifa, cho2023davidsonian, cho2023visual}. Each \metric relies on an external language model (LM) to generate questions given the text prompt. %These metrics rely on two external model components which feed the consistency scoring \citep{hu2023tifa, cho2023davidsonian, cho2023visual}: (1) a language model (LM) and (2) a visual question answering (VQA) model. First, the text prompt is fed into the LM, which is instructed to generate several questions from it:
In the simplest case, the LM might generate questions: ``\textit{is there a dog?}'', ``\textit{is the dog purple?}'', ``\textit{are there flowers?}'', etc.
% Then, each LM-generated question is paired with the T2I model-generated image, in turn, and fed into the VQA model, which will output an answer (for example, ``yes'', ``yes'', and ``yes'', if the original prompt and the generated image were consistent). An overall consistency score can then be calculated by averaging the number of correct answers the VQA model gives. 
Then, these LM-generated questions are passed to computer vision (CV) models, typically visual question answering (VQA) models, which calculate an overall consistency score by averaging the correct answers to the questions given the image.

Because these recent automatic scoring approaches rely on the simplicity of LMs and the interpretability of CV modules like VQA, they are being increasingly adopted, with new variations on previous metrics being proposed at a rapid pace. In this work, we take stock of where we are, and determine which (if any) of the existing metrics are most informative. To do this, we take a step back and assemble a list of very basic desiderata that an ideal automatic metric for T2I consistency should be expected to satisfy; see Table \ref{tab:criteria}. Next, given our set of desiderata, we evaluate four existing \metrics to see if they satisfy these ideal properties and find that none of the tested \metrics actually satisfy all of them.

\newcommand{\tableCellLabel}[1]{\parbox{6em}{\small\centering #1}}

\begin{table*}[t!]
    \centering
    %Desiderata for Strong Evaluation Metrics\\[2ex]
    \small
    % removed "correlates with human judgement"
    \begin{tabular}{c|cccccc}
        %& \multicolumn{5}{c}{placeholder}
        % & \multicolumn{2}{c}{placeholder} \\

        % \toprule

        \parbox{4em}{Evaluation\\Metric}
        & \parbox{4.5em}{\centering Human\\Interpretable}
        & \parbox{5em}{\centering External\\Models\\(Section \ref{sec:approach})}
        & \parbox{7em}{\small\centering Sensitive\\to Text Properties\\(Section \ref{sec:corr-ling}, \ref{sec:ablations})}
        & \parbox{7em}{\centering Sensitive\\to Image Properties\\(Section \ref{sec:corr-visual}, \ref{sec:ablations})}
        % & \tableCellLabel{Contributes \\Meaningful\\Additional Info\\(Section \ref{sec:metrics-distinct})}
        % & \tableCellLabel{Suitably Baselined\\with\\Ablations\\(Section \ref{sec:ablations})}
        & \parbox{5.5em}{\centering Robust to\\ Known Shortcuts (Section \ref{sec:textartifact})}
         \\\toprule
         
         % \small Humans
         %& \cmark & N/A & \cmark & $\sim$ & N/A & N/A  \\\midrule
         
         \small CLIPScore
         & \red\xmark & CLIP & \green\cmark & \orange$\sim$ &  N/A \\\midrule
         
         \small TIFA
         & \green\cmark& LM, VQA & \green\cmark & \orange$\sim$ &  \red\xmark \\\midrule
         
         \small VPEval
         & \green\cmark & \parbox{6em}{\centering LM, VQA,\\obj. detector, OCR}  & \green \cmark & \orange$\sim$ &   \red\xmark \\\midrule
         
         \small DSG & \green\cmark & LM, VQA & \green\cmark & \orange$\sim$ &   \red\xmark\\

         % \small FID &  &  & & & feature extractor & & \\

         \bottomrule
    \end{tabular}
    \\[1.5ex]
    \scalebox{0.9}{LM = language model \hspace*{1.5em}
    {VQA = visual question answering model \hspace*{1.5em} 
    OCR = optical character recognition}}
    \caption{
        Desiderata for strong and informative \metrics for text-to-image models. Criteria with mixed signal are marked with ``{\orange$\sim$}".
        %\\[3ex]ablations done in the paper for each approach -- CLIPScore: FOIL dataset to test sensitivity to false captions
        %\TODO{\textit{minor:} align the text in top column}
    }
    \label{tab:criteria}
\end{table*}

We additionally explore the relationship \textit{between} existing metrics and find correlations with CLIPScore are low, suggesting the new \metrics may genuinely contribute novel information above and beyond CLIPScore. However, we also measure how well metrics that were proposed earlier correlate with those that were proposed later, such as TIFA for VPEval and DSG, and VPEval for DSG, and find that all three correlate with each other to a medium or strong degree.

We also perform some ablations to better understand how much text and image information are leveraged. Our results provide additional evidence that all \metrics have serious weaknesses in that they insufficiently rely on visual information, and also questions are raised about their text abilities as well. 

Our results suggest there is ample room to further refine and extend our existing suite of automatic \metricsNoSpace. Until we have a firm idea of what it is that we want our metrics to accomplish, it will continue to be challenging to design adequate metrics. Our work has taken some initial steps towards proposing a handful of minimal desiderata, but future work could also incorporate additional desiderata to help guide the design of better automatic \metrics that are more robust and can better evaluate the performance of text-to-image generation models.

\section{Approach}\label{sec:approach}
\subsection{What makes a good \metricNoSpace ?}

Metric conceptualization and operationalization have long been a core part of the scientific work of evaluation in ML research fields \citep{graham-2015-evaluating, welty2019metrology, jacobs2021measurement}. In NLP and in CV, such work focuses on everything from designing metrics that better measure their underlying constructs \citep{howcroft-etal-2020-twenty, blodgett-etal-2021-stereotyping, kiela2021dynabench, xiao-etal-2023-evaluating-evaluation}, to understanding metric correlations \citep{liu-etal-2023-question, sun-etal-2023-validity}, from taking into account relevant control experiments \citep{barbu2019objectnet} to devising better evaluations and evaluation metrics that avoid shortcuts \citep{geirhos2020shortcut} and other features that may make measurement unreliable or hard to interpret \citep{jia-liang-2017-adversarial, gururangan-etal-2018-annotation, tu-etal-2020-empirical, blodgett-etal-2021-stereotyping, raji-etal-2021-everything, 
wang-etal-2022-identifying, banerjee2023shortcuts, cummings2023objectnet, sinha-etal-2023-language, zheng2023large}.

In this work, we focus on the construct validity of four existing automatic text-to-image (T2I) consistency metrics, all of which were recently proposed.  As a first stab, all four metrics could arguably be deemed construct valid, as they have all been demonstrated to correlate highly with human judgements, and have additional desirable properties, such as human interpretability. However, we argue there are several additional properties strong T2I \metrics should have (see \autoref{tab:criteria} for our minimal desiderata), and none of our investigated metrics have all of them. 

In general, desiderata for metrics fall into three classes: (i) desiderata that are necessary for every evaluation metric, (ii) desiderata that are necessary when proposing new evaluation metrics, and finally, (iii) nice-to-haves. For our necessary criteria, a \metric should be sensitive to images \citep{antol2015vqa} and sensitive to text, if it is to measure the consistency between the two (see Section~\ref{sec:corr-ling}, Section~\ref{sec:corr-visual} and Section~\ref{sec:ablations}). It should also actually measure \faithfulness in a way that is not affected by previously identified, unwanted artifacts or shortcuts (see Section~\ref{sec:textartifact}). 

For necessities when proposing a new metric, newly proposed metrics should also improve above and beyond reasonable baselines, including random baselines, and outperform existing alternative metrics, such as CLIPScore (see \autoref{tab:overall-metric-results}). Another desiderata for proposing a new metric is showing that it contributes additional important information, understanding, or contextualization that the previous metric(s) lacked---to explore this point, we also measure how strongly the three \metrics correlate with each other and with an existing T2I metric, CLIPScore (see Section~\ref{sec:metrics-distinct}).

Of course, the few desiderata we explore here are not intended to be fully exhaustive. They are intended to be a starting point, a minimal set of properties that automatic \metrics should have. We discuss other desiderata that we might also want our \metrics to satisfy in Section~\ref{sec:discussion} below.

\subsection{Evaluation Metrics -- Text-Image Consistency}\label{sec:eval-metrics}

We focus on four metrics -- %for evaluating \faithfulness --
CLIPScore, TIFA, VPEval and Davidsonian Scene Graph (DSG). 

\paragraph{CLIPScore.}
% \TODO{} decide whether we want to use the weight multiplier

CLIP \citep{radford2021learning} is a vision-language model that maps images and text to a feature embedding space.
%It has a text encoder and an image encoder to extract features $f_{\textrm{text}}$ and $f_{\textrm{image}}$ from each modality. 
CLIPScore \citep{hessel2021clipscore} approximates the text-image consistency by using the cosine similarity between the features of the image and the text using CLIP.

\paragraph{TIFA.}

TIFA, or Text-to-Image Faithfulness Evaluation \citep{hu2023tifa}, uses two primary components -- an LM to generate questions from the text prompt and a VQA model to answer the questions using the generated images. The score is computed as the percent of correctly answered questions from the VQA model.

\paragraph{VPEval.} VPEval \citep{cho2023visual} generates visual programs from the text prompt using an LM. Where TIFA questions are in natural language % (\eg \texttt{Are there 2 cats? yes")}), VPEval instead generates visual programs (\eg \texttt{count("cat", ==2}).
The visual programs are executed by 8 modules, such as
%object detection, object evaluation, counting, text evaluation, spatial evaluation, scale evaluation, optical character recognition (OCR) and VQA --
scale evaluation and counting, that use 3 different vision and vision-language modules including an object detector, OCR model and VQA model.% The object detector module can be evaluated using either an explicit object detector \textit{or} using a VQA model.

\paragraph{Davidsonian Scene Graph (DSG).} Davidsonian Scene Graph (DSG) \citep{cho2023davidsonian} is similar to TIFA, using LM-generated questions and a VQA model. The key difference is that DSG focuses on addressing inconsistent and hallucinated answers. Questions are counted as correct if and only previous dependencies were also correct. For instance, if the VQA model incorrectly predicts there is not a dog while correctly predicts that the dog is red in the second question, the second question about the red dog is marked incorrect.

\subsection{Text-to-Image (T2I) Models}
We use three state-of-the-art T2I models: (i) a diffusion model that leverages CLIP embeddings via a paid API (\texttt{\dmwithclipNoSpace}), (ii) a latent diffusion model (LDM) for which we evaluate two checkpoints --  \texttt{\ldmOneNoSpace} (from a paid API) and \texttt{\ldmTwoNoSpace} (open-sourced) -- and (iii) the open-source version of \texttt{\glideNoSpace}.

\section{Experiments}

\begin{table*}[!ht]
\centering
\scalebox{0.95}{
    \begin{tabular}{l|llll|llll}
    \toprule
    \multicolumn{1}{c|}{}
    & 
    \multicolumn{4}{c|}{\textbf{COCO}} &
    \multicolumn{4}{c}{\textbf{Winoground}} \\
    T2I Source & CLIP & TIFA & VPEval & DSG 
    & CLIP & TIFA & VPEval & DSG \\ \midrule

    Random Chance & N/A & 43.1 & 39.1 & 33.4
    & N/A & 44.0 & 40.6 & 36.0
    \\
    
    % Text-only QA & N/A & 60.0 & 54.3 & 76.2
    % & N/A & 58.8 & 51.6 & 67.2 \\

    Real Images & 30.6 & 83.0 & 77.3 & 79.2
    & \bf 21.5 & 66.8 & 60.8 & 63.7 \\ \hdashline
    
    \glide & \bf 28.9 & \bf 64.7 & \bf 59.6 & \bf 55.4
    & \bf 21.5 & \bf 51.1 & \bf 45.6 & \bf 41.5 \\
    
    % \glidebowshort & 28.1 & 60.9 & 56.8 & 49.6
    % & 21.2 & 50.5 & 45.0 & 40.6 \\
    
    \dmwithclipshort & \bf 31.7 & \bf 85.3 & \bf 79.1 & \bf 81.5
    & \bf 25.0 & \bf 71.0 & \bf 64.6 & \bf 68.4 \\

    \ldmOne & 31.2 & 79.2 & 73.3 & 73.9 
    & 24.4 & 66.3 & 59.4 & 60.5 \\
    
    \ldmTwo & 31.0 & 79.1 & 73.4 & 73.6
    & 23.5 & 62.7 & 57.5 & 58.1 \\
        
     \bottomrule
    \end{tabular}
}
\caption{Results for 4 \metrics evaluated on four T2I models, as well as a random chance baseline.
While our analysis is not focused on relative performance between different T2I models, we do \textbf{bold} the highest and lowest performing models for each \metricNoSpace. % 
}
\label{tab:overall-metric-results}
\end{table*}

Before our deeper analysis, we first report the \metric scores for five T2I models in \autoref{tab:overall-metric-results}. We use the datasets MS-COCO \citep{lin2014microsoft} and Winoground \citep{thrush2022winoground} for text prompts. In total, we have 11,525 and 769 generated images per model for COCO and Winoground, respectively.\footnote{We omit all prompts that refer to people.}

For evaluating CLIPScore, we use the CLIP ViT-L14 checkpoint provided by OpenCLIP \citep{ilharco_gabriel_2021_5143773}. For TIFA, VPEval and DSG, we generate questions using Llama-v2-Chat 70B checkpoint model \citep{touvron2023llama}.
For ease of comparison, we use the newer BLIP2-Flan T5 XL \citep{li2023blip} for all VQA questions. For the remaining, non-VQA questions in VPEval, we use the same models from their paper. For simplicity, we refer to TIFA, VPEval and DSG as \textbf{\textit{VQA-based metrics}}.

All metrics exceed a na\"\i ve random chance baseline.
The \metrics rank models the same on COCO and Winoground (\texttt{\dmwithclipshort $>$ \ldmOne $>$ \ldmTwo $>$ GLIDE}). This consistency suggests either that all metrics are differently informative and their agreement reflects genuine T2I model ranking, or that all metrics actually contribute the same kind of information and are redundant. Section~\ref{sec:metrics-distinct} below will adduce evidence for the latter interpretation. We also observe that, for all metrics, the generated images from at least one T2I model actually score higher than the real images. This could be because real images are visually richer, denser scenes with more necessary information to process.

\subsection{Experiment 1: Relationship with linguistic properties}\label{sec:corr-ling}

An ideal \metric should draw on information from provided text \textit{and} from the image. First, to determine whether the \metrics are highly dependent on the text in particular, we evaluate the correlation between the metrics and several standard linguistic properties of the text prompt. We measure Spearman's rank correlation between the \metric and the \textbf{\textit{readability}}, \textbf{\textit{complexity}} and \textbf{\textit{length}} of the prompt. For readability, we use the Flesch–Kincaid grade level calculation \citep{flesch1948new}. Flesch-Kincaid approximates the difficulty of reading a passage, based in part on word and sentence length, with higher scores corresponding to more difficulty. For complexity, we use Yngve scores \citep{yngve1960model}, which use constituency parsing. Deeper and wider of parse trees means more complex sentences. Finally, for prompt length, we use NLTK's word tokenizer \citep{loper2002nltk} to get a word count, excluding stopwords.

Results are shown in Table \ref{tab:linguistic-corr-coco} for COCO, and in Table \ref{tab:ling-corr-winoground}  in the Appendix for Winoground. Overall, we find that all four metrics on all models are correlated to a medium-to-strong degree with all linguistic properties for COCO ($-0.4$ to $-0.94$) and with the length property for Winoground. This suggests that these metrics are sensitive to the linguistic properties of the text prompts. In general, VPEval and DSG significantly correlate with nearly all linguistic properties in COCO for nearly all models, whereas TIFA and CLIPScore show weaker effects.  VQA-based metrics  \textit{negatively} correlate with the linguistic properties, probably because ``harder" prompts (longer, more complex, higher grade-level) can solicit lower \metric scores.\footnote{Prompt difficulty affect one  components in the evaluation pipeline, or it may cascade. Perhaps the T2I model has trouble generating images from harder prompts, leading to low scores, and a negative correlation. Alternatively,  perhaps the LM struggles to questionize hard prompts and/or the VQA model struggles to answer them. For our purposes, the existence of these correlations is sufficient to motivate our conclusions, although future work could try to isolate which subcomponents suffer from insensitivity to visual and/or textual information of \metricsNoSpace.}  CLIPScore is not particularly sensitive to syntactic complexity, supporting prior work that CLIP operates more as a bag of words \citep{yuksekgonul2022and}; this may also explain why it positively correlates with grade-level and prompt length.

\newcommand{\tableCellLing}[1]{\rotatebox{80}{\parbox{4em}{ #1}}}
\newcommand{\tableHeaderLing}[1]{\parbox{13em}{\centering #1 \vspace*{1.5ex}}}
\newcommand{\tableCellVisual}[1]{\rotatebox{80}{\parbox{4em}{ #1}}}

\begin{table*}[!ht]
    \centering
    %{\small Spearman's Rank Correlation $\rho$ between Metrics and Linguistic Properties for COCO}\\[1.5ex]
    \begin{subtable}{\textwidth}
    \resizebox{\textwidth}{!}{
    \begin{tabular}{l|llll |llll| llll}
    & \multicolumn{4}{c}{
        \tableHeaderLing{$\rho$ -- Readability (Grade Level)}
        }
    & \multicolumn{4}{c}{
        \tableHeaderLing{$\rho$ -- Syntactic Complexity}
        }
    & \multicolumn{4}{c}{
        \tableHeaderLing{$\rho$ -- Length (\# of Words)}
        } \\

    & \tableCellLing{CLIPScore}
    & \tableCellLing{TIFA}
    & \tableCellLing{VPEval}
    & \tableCellLing{DSG}
    
    & \tableCellLing{CLIPScore}
    & \tableCellLing{TIFA}
    & \tableCellLing{VPEval}
    & \tableCellLing{DSG}
    
    & \tableCellLing{CLIPScore}
    & \tableCellLing{TIFA}
    & \tableCellLing{VPEval}
    & \tableCellLing{DSG}
    \\\midrule
    
    %\hline
    Real Images & 0.28* & -0.30* & \bf -0.54* & \bf -0.44* & 0.23* & -0.36* & \bf -0.46* & \bf -0.44* & 0.29 & \bf -0.45* & \bf -0.76* & \bf -0.74*  \\
    
    \glide & -0.04 & -0.31* & \bf -0.63* & \bf -0.44* & -0.01 & -0.37* & \bf -0.41* & \bf -0.41* & -0.10 & \bf -0.66* & \bf -0.72* & \bf -0.77*  \\
    
    %glide\_bow & 0.19 & -0.38* & -0.45* & -0.38* & 0.08 & -0.30* & -0.37* & -0.29* & 0.39* & -0.59* & -0.69* & -0.76* & -0.93* & -0.93* & -0.93* & ~ \\
    
    \dmwithclipshort & 0.33* & -0.22 & \bf -0.66* & \bf -0.40* & 0.15* & -0.39* & \bf -0.45* & \bf -0.48* & 0.28 & \bf -0.70* & \bf -0.80* & \bf -0.94*  \\
    
    \ldmOne & \bf 0.49* & \bf -0.42* & \bf -0.56* & -0.30* & 0.10 & -0.34* & \bf -0.51* & \bf -0.48* & \bf 0.50* & \bf -0.74* & \bf -0.86* & \bf -0.91*  \\

    \ldmTwo & \bf 0.41* & -0.35* & -0.38* & -0.38* & 0.04 & -0.36* & \bf -0.50* & \bf -0.46* & \bf 0.42* & \bf -0.66* & \bf -0.69* & \bf -0.84*  \\\bottomrule
    
    \end{tabular}
    }
    \caption{Spearman's rank correlation between \textit{linguistic properties} and \metrics for COCO. Metrics are highly sensitive to linguistic features of the text. Statistically significant values are marked with $*$. Moderate to strong statistically significant correlations are \textbf{in bold}. %See Table \ref{tab:ling-corr-winoground} for Winoground results.
    }
    \label{tab:linguistic-corr-coco}
    \end{subtable}
    
    \centering
    %{\small Spearman's Rank Correlation $\rho$ between Metrics and Visual Properties for COCO} \\

    \begin{subtable}{\textwidth}
    \resizebox{\textwidth}{!}{
    \begin{tabular}{l| llll| llll| llll}
    
    \multicolumn{1}{c}{}
    & \multicolumn{4}{c}{$\rho$ -- Concreteness}
    & \multicolumn{4}{c}{$\rho$ -- Imageability}
    & \multicolumn{4}{c}{\parbox{8em}{$\rho$ -- ImageNet-21k\\Caption Overlap}}\\

    & \tableCellVisual{CLIPScore}
    & \tableCellVisual{TIFA}
    & \tableCellVisual{VPEval}
    & \tableCellVisual{DSG}
    
    & \tableCellVisual{CLIPScore}
    & \tableCellVisual{TIFA}
    & \tableCellVisual{VPEval}
    & \tableCellVisual{DSG}
    
    & \tableCellVisual{CLIPScore}
    & \tableCellVisual{TIFA}
    & \tableCellVisual{VPEval}
    & \tableCellVisual{DSG}\\
    \toprule
    
    Real Images & 0.04* & -0.03 & 0.10* & -0.02 & -0.06* & 0.00 & 0.08* & 0.00 & -0.05 & -0.03 & 0.23 & 0.0 \\
    
    \glide & 0.03* & -0.05* & 0.00 & -0.04* & 0.12* & 0.05* & 0.10* & 0.06* & -0.12 & -0.08 & -0.05 & -0.07 \\ 
    
    % glide\_bow & 0.05* & -0.05* & -0.00 & -0.04* & 0.10* & 0.03* & 0.08* & 0.02 & 0.07 & -0.05 & 0.07 & -0.04 \\
    
    \dmwithclipshort & 0.08* & -0.02 & 0.09* & -0.01 & -0.01 & 0.02 & 0.09* & 0.03* & 0.06 & -0.06 &  0.24 & 0.08 \\

    \ldmOne & 0.02 & -0.03* & 0.09* & -0.03* & 0.02 & 0.02 & 0.10* & 0.02 & -0.04 & -0.04 & 0.23 & 0.11 \\
    
    \ldmTwo & 0.02 & -0.04* & 0.06* & -0.05* & 0.04* & 0.01 & 0.09* & 0.01 & 0.04 & -0.17 & -0.01 & 0.02 \\
    
    \bottomrule
    \end{tabular}
    }
    \caption{Spearman's rank correlation between \textit{visual properties} and \metrics. Metrics are \textit{not} sensitive to visual properties we evaluated. Statistically significant values are marked with $*$. 
    }
    \label{tab:visual-corr-coco}
    \end{subtable}
    
    \caption{Correlation for COCO between the \metrics and \textit{linguistic properties} are generally moderate to strong across models, while the correlation between these metrics and \textit{visual properties} are predominantly weak. These results suggest \metrics are more language-related than vision related.% We do not find as strong correlations for visual features; see Table \ref{tab:visual-corr-coco} in the Appendix. 
    % We also observe similar patterns for the Winoground dataset; results are shown in Table \ref{tab:ling-corr-winoground} in the Appendix.
    %\textbf{Takeaway:} \Metrics are sensitive to linguistic properties of the text prompt, showing moderate to strong correlation. However, these properties do not appear to be sensitive to visual properties.
    }
    \label{tab:corr-coco}
\end{table*}

\subsection{Experiment 2: Relationship with visual properties}\label{sec:corr-visual}

Next, we analyzed the metrics' relationship with visual features. %, as proxied by the text prompt.
These include the \textbf{\textit{imageability}} (how easily hearing a word leads to creating a mental image, \citealt{paivio1968concreteness,bird2001age}), \textbf{\textit{concreteness}} (how easily a word can be experiences by the senses,  \citealt{paivio1968concreteness}) and \textbf{\textit{overlap with ImageNet-21k (IN-21k) object classes}} \citep{ridnik2021imagenet}. We use imageability ratings from \citet{gao2023scope} and concreteness ratings provided by \citet{brysbaert2014concreteness} and average across the words in the sentence \footnote{For concreteness and imageability, we assign any missing words the lowest imageability/concreteness score in the corpus. We also tested out (i) assigning the missing words a score of 0 and (ii) omitting missing words. We observed no discernible difference in the correlations.}. For IN-21k, we compute the percentage of words in the prompt that are also IN-21k objects classes. For all visual property calculations, we exclude stopwords. 
See Table \ref{tab:visual-corr-coco} for COCO results, and \ref{tab:visual-corr-winoground} in the Appendix for Winoground results. 

Because language provides such a strong prior in vision-language tasks like VQA \citep{zhang2016yin,goyal2017making,agrawal2018CVPR, lin2024revisiting}, we wanted to ensure the visual modality was being used in these metrics. We found essentially no correlation between the \metrics and the visual properties we evaluated, suggesting the metrics may insufficiently leverage visual properties.

\if0
\paragraph{Do evaluation metrics correlate with prompt concretetess?}\label{sec:concreteness} 
Are prompts that describe more concrete words and concepts more accurately generated? We use the concreteness ratings for 40k words provided by \citet{brysbaert2014concreteness} to approximate a concreteness score for each prompt. We use the lowest concreteness score in the corpus for any missing words.

\paragraph{Do evaluation metrics correlation with prompt imageability?}
Imageability describes how easily hearing a word leads to creating a mental image \citep{paivio1968concreteness,bird2001age}. We use imageability scores from \citet{gao2023scope}. We follow the same approach as we did in Section \ref{sec:concreteness} to handle prompts that contain words without an associated imageability score. While we expected that more imageable prompts would correlate with higher \metrics scores, we did not observe this.

\paragraph{Do evaluation metrics correlate with the number of objects in an image?}

ImageNet21k (IN21k; \citealt{deng2009imagenet}) is a dataset of 14M images and 21k different labels. This dataset has been used for pretraining and finetuning many vision models (\eg \citealt{steiner2021train,zhou2022detecting}). For this reason, we explore whether captions that have a high overlap with classes in IN21k correlate with higher performance. We hypothesize that text prompts that include IN21k should be easier for T2I models to generate and should therefore have higher \metrics scores.
\fi

\subsection{Experiment 3: How distinct are these metrics?}\label{sec:metrics-distinct}

\begin{figure*}[hb!]
    \centering
    \hspace*{-1em}
    \begin{tikzpicture}
    % top row
    \node(real-images){
        \includegraphics[width=10.5em,trim={1em 0 0  0},clip]{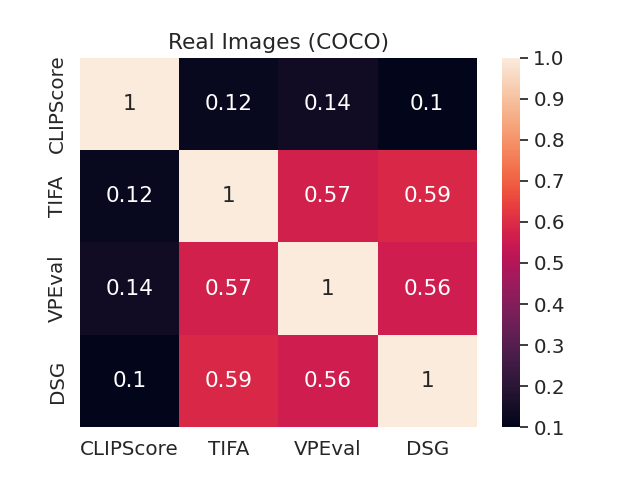}
    };
    \node[right=-3.5em of real-images](dmwithclip2){
        \includegraphics[width=10.5em,trim={3em 0 0 0},clip]{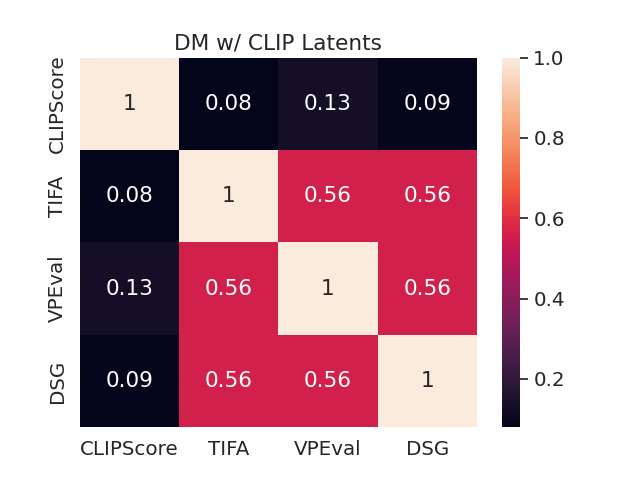}
    };
    \node[right=-3.5em of dmwithclip2](glide){
        \includegraphics[width=10.5em,trim={3em 0 0 0},clip]{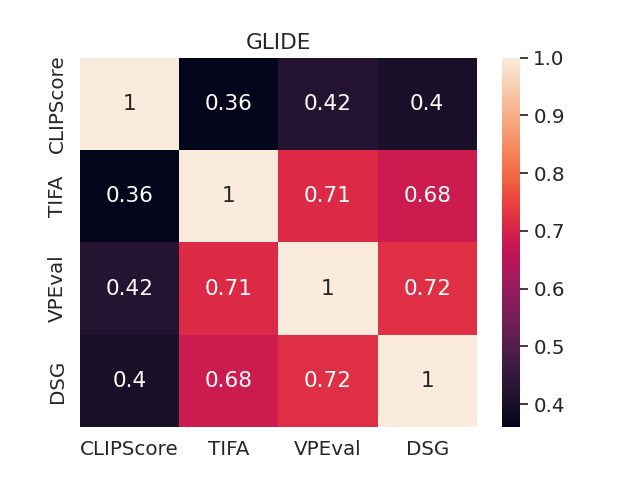}
    };
    \node[right=-3.5em of glide](sd-hf){
        \includegraphics[width=10.5em,trim={3em 0 0 0},clip]{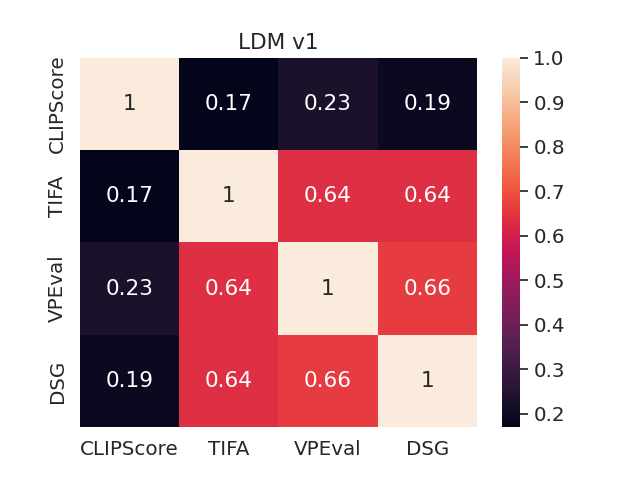}
    };
    \node[right=-3.5em of sd-hf](sd-api){
        \includegraphics[width=10.5em,trim={3em 0 0 0},clip]{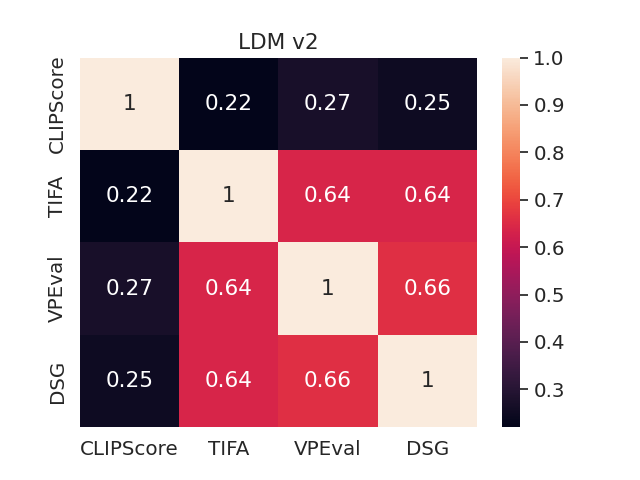}
    };
    \end{tikzpicture}
    
    \caption{Correlation between each pair of \metrics for COCO. The VQA-based metrics are not highly correlated with CLIPScore (excluding \glide), and correlations with real images resemble those from generated images. For VQA-based metrics, correlations are medium to strong and statistically significant, suggesting they may be interchangeable. We observe broadly similar trends for Winoground; see Figure \ref{fig:exp3-winoground}. % \textbf{Takeaway: The VQA-based \metrics have moderate and strong correlations for real and generated images.}
    }
    \label{fig:exp3-coco}
\end{figure*}

When a new evaluation metric is proposed, we should first check to be sure that it pushes the state of the art. In other words, it should probe new information (or probe old information in a better way). We investigate the extent to which newer VQA-based metrics convey similar information to existing metrics (CLIPScore) and to each other, quantifying similarity as Spearman's rank correlation. We use Spearman's, because it is a nonparametric test and doesn't make strong assumptions about the shape of the underlying distribution. We generate images for every prompt in %$\mathcal{D}_{COCO}$ and $\mathcal{D}_{Winoground}$ 
COCO and Winoground across the 5 different T2I models and using the real images. Then, we compute the scores for metrics (CLIPScore, TIFA, VPEval and DSG) and then measure the pairwise Spearman's correlation between the metrics. Results are shown in Figure \ref{fig:exp3-coco}.

\textbf{VQA-based metrics are strongly and significantly correlated with each other.} This may be due to either 1) similar approaches to generating questions using an LM or 2) similar approaches to evaluation by using VQA. CLIPScore shows the weakest correlations with other metrics, perhaps because it operates using cosine similarity. This leads us to question -- how can we ensure new VQA-based evaluation metrics introduce or leverage new information?

\subsection{Experiment 4: Zooming in on the Generated Questions}\label{sec:textartifact}

Given that our results up until this point show  text information matters for the metrics, we decided to foreground text-based analyses of the questions for the rest of this section. 

\paragraph{Number of generated questions correlates strongly with \metricsNoSpace.}

For this experiment, we analyzed the number of questions generated by the LM. We omit CLIPScore, which uses the caption directly. We find that the number of questions negatively correlates with metric scores, especially for COCO (see \autoref{ref:tab-num-questions}). This behavior makes some sense, because including more questions gives the model more chances to make a mistake.

\begin{table}[!ht]
\centering
%{\small Spearman's Rank Correlation $\rho$ between Metrics and Linguistic Properties for COCO}\\[1.5ex]
    
    \begin{tabular}{l| lll | lll}
    & \multicolumn{3}{c}{\it COCO}
    & \multicolumn{3}{c}{\it Winoground} \\
    
    & TIFA
    & VPEval
    & DSG
    
    & TIFA
    & VPEval
    & DSG \\\midrule
    
    %\hline
    Real Images 
    & \bf -0.93* & \bf -0.93* & \bf -0.93*
    & \bf-0.32 & 0.16 & \bf-0.54 \\
    
    \glide 
    & \bf -0.93* & \bf -0.93* & \bf -0.93* 
    & -0.24 & \bf -0.44 & \bf -0.64* \\
    
    %glide\_bow & 0.19 & -0.38* & -0.45* & -0.38* & 0.08 & -0.30* & -0.37* & -0.29* & 0.39* & -0.59* & -0.69* & -0.76* & -0.93* & -0.93* & -0.93* & ~ \\
    
    \dmwithclipshort
    & \bf -0.92* & \bf -0.92* & \bf -0.92*
    & -0.08 & -0.27 & \bf -0.84* \\
    
    \ldmOne
    & \bf -0.97* & \bf -0.97* & \bf -0.97* 
    & \bf -0.44 & -0.17 & \bf -0.85* \\

    \ldmTwo
    & \bf -0.96* & \bf -0.96* & \bf -0.96*  & -0.35 & -0.20 & -0.31 \\
    \bottomrule
    
\end{tabular}
\caption{Spearman's correlation $\rho$ between \# of generated questions and the \metrics are negative and very large (larger than $-0.9$ for every metric on COCO). 
}
\label{ref:tab-num-questions}
\end{table}

However, if the correlation between our metrics and basic properties of one of its subcomponents, namely the LM, is strong, we might ask whether having the whole evaluation pipeline genuinely contributes more than just using the LM. In this case, the answer seems fairly clear that the LM may be sufficient, with high significant correlations of above $-0.9$ for all models on COCO.\footnote{Given this strong correlation, we were curious to learn more about this relationship. Visually and mathematically (with a Pearson's correlation that was insignificant), we confirmed that it is not linear, but more research is necessary to fully characterize it. 
} While this result is especially stark for COCO, the same trend holds on Winoground, which is a more challenging but smaller dataset, for DSG, but not as strongly for TIFA and VPEval. These results seem to suggest that the LM question generation stage alone may supply enough signal to infer the \metric scores. This would imply that the VQA component can be omitted from the evaluation pipeline.

\paragraph{Distribution of VQA Questions.}

Next, we aim to determine whether the \metrics might be relying on shortcuts, and/or falling prey to unwanted statistical artifacts. First, recall that \metrics pipeline relies on two models: (i) an LM which takes in the prompt and outputs a set of questions and their ``ground truth'' answers, and (ii) a VQA model that takes in the LM-generated questions and the image generated by the T2I model, and generates answers. The VQA model's answers are then matched to the LM's ``ground truth'' answers to get a score.  Recall that the LM is prompted, by design, to generate binary questions with ``yes'' answers, or, for TIFA and VPEval, 4-option multiple choice questions with first-correct answers (see \autoref{tab:qas-stats}). 

Yes- or first-correct ground truth answers is problematic when the pipeline includes a VQA model. Since \citet{antol2015vqa} originally proposed the VQA as a task, statistical biases in VQA have been a major topic of research \citep{agrawal2018CVPR, ray2019sunny, shah2019cycle, agarwal2020towards,sheng2021human}. It's particularly salient in the field that VQA models suffer from \texttt{yes}-bias \citep{zhang2016yin} and \texttt{first-answer}-bias, meaning they output these two answers at very high base rates (LMs exhibit similar problems \citep{benchekroun2023worldsense,zheng2023large}). This means that the na\"\i ve random chance baseline we reported in \autoref{tab:overall-metric-results} may not actually be at all informative -- we need a majority class baseline for the VQA model. Absent that, we genuinely cannot be sure whether a ``yes'' (or first-correct) output from the \metrics pipelines means that the text and image are genuinely consistent, or whether it just means that the VQA model is spuriously generating according to its skewed distribution regardless of the inputs. This fact has two additional implications: (i) a ``no'' (or non-first answer) will be strictly more informative, and (ii) a program of a few lines that merely prints ``yes'' (or the first answer) could replace the VQA component entirely, and still yield high \metric scores without ever seeing any image or prompt. 

A yes- or first-correct only evaluation setup not only benefits from VQA artifacts, it also represents a break from more classical VQA evaluation \citep{antol2015vqa, zhang2016yin}, where the distribution of ground truth test answers is assumed to match the distribution of the training data (no VQA model, to our knowledge has been purposefully trained to always output ``yes'', even if they do output it at high rates). One could return to the classical testing set up, whereby the test distribution reflects the underlying -- albeit ``yes''-skewed, distribution of the VQA system -- prompting the LM to generate questions where the ground truth answer should be ``no''. One could also draw on the classical ML research on class imbalance \citep{he2013imbalanced,fernandez2018learning,henning-etal-2023-survey}, taking answer skew into account when calculating final accuracy, perhaps by resampling answer classes to match the VQA training distribution \citep{buda2018systematic}.

\begin{figure*}[ht!]
    \centering
    \begin{subfigure}{0.9\textwidth}
        \includegraphics[width=\textwidth]{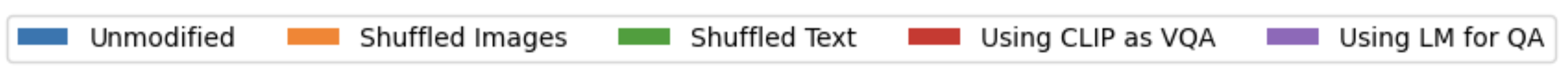}
    \end{subfigure}

    \hspace*{-3em}
    \begin{subfigure}{0.46\textwidth}
        \includegraphics[width=1.1\textwidth]{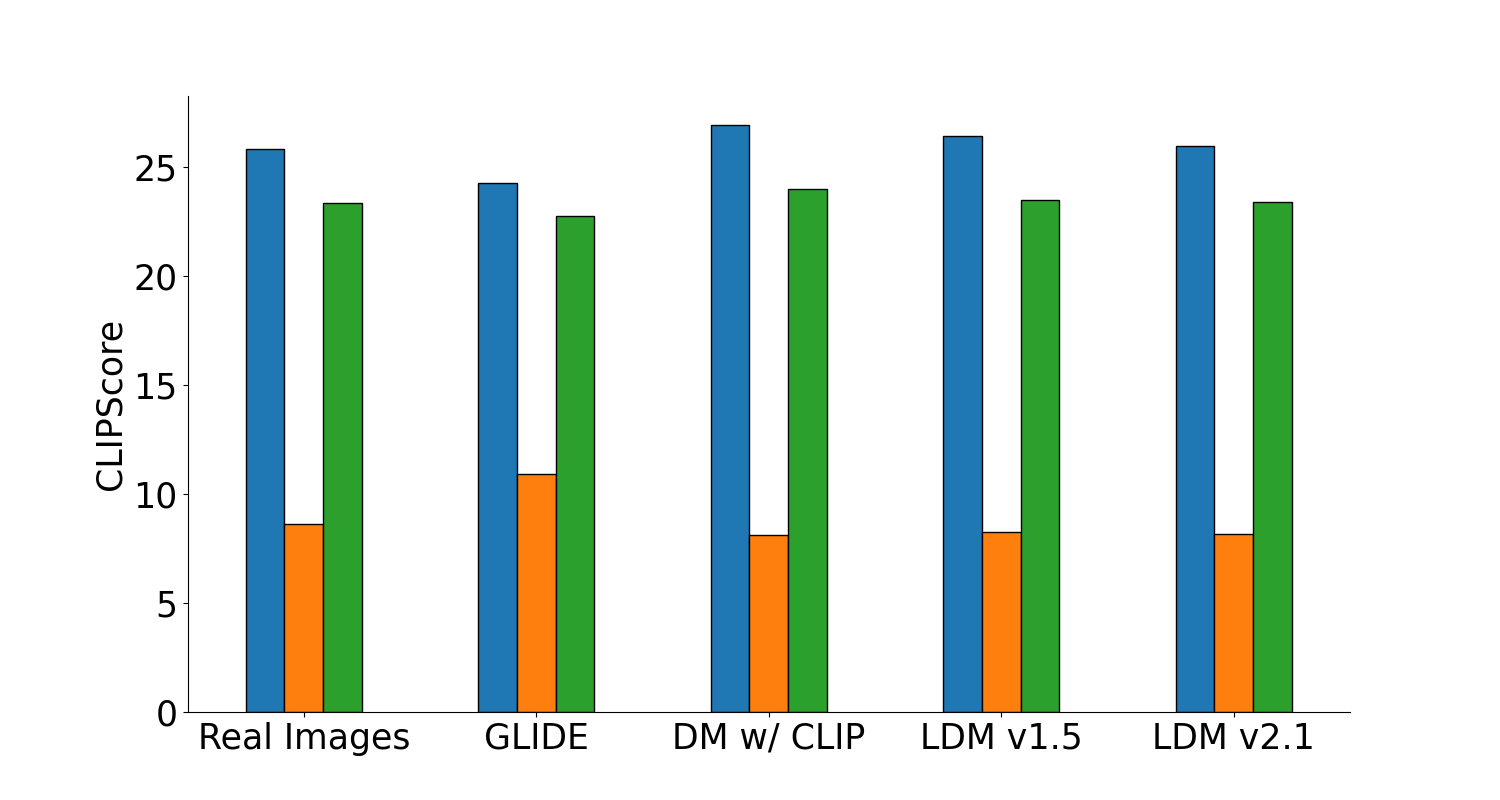}
    \end{subfigure}
    \hspace*{0em}
    \begin{subfigure}{0.46\textwidth}
        \includegraphics[width=1.1\textwidth]{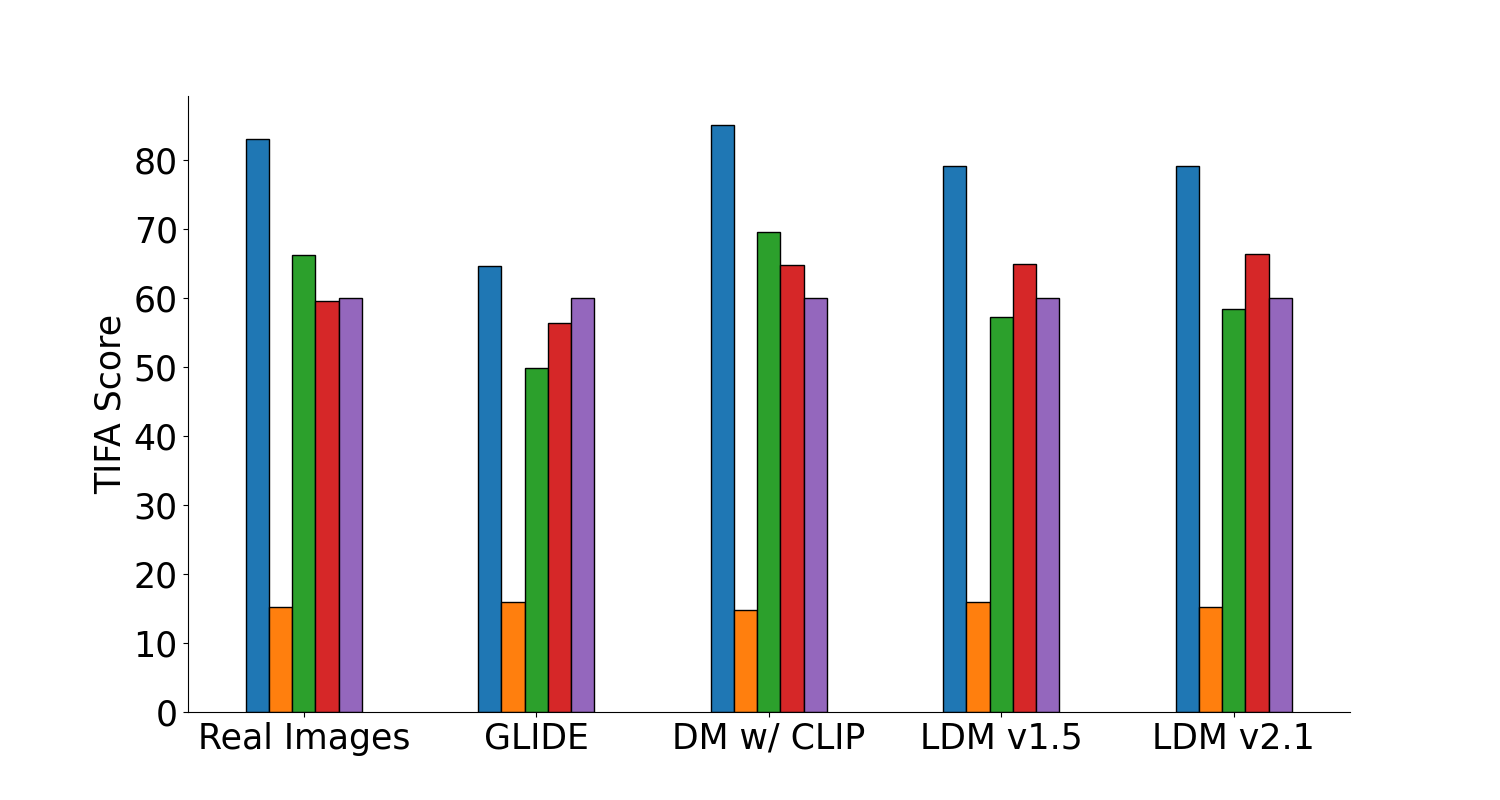}
    \end{subfigure}
    \\

    \hspace*{-3em}
    \begin{subfigure}{0.46\textwidth}
        \includegraphics[width=1.1\textwidth]{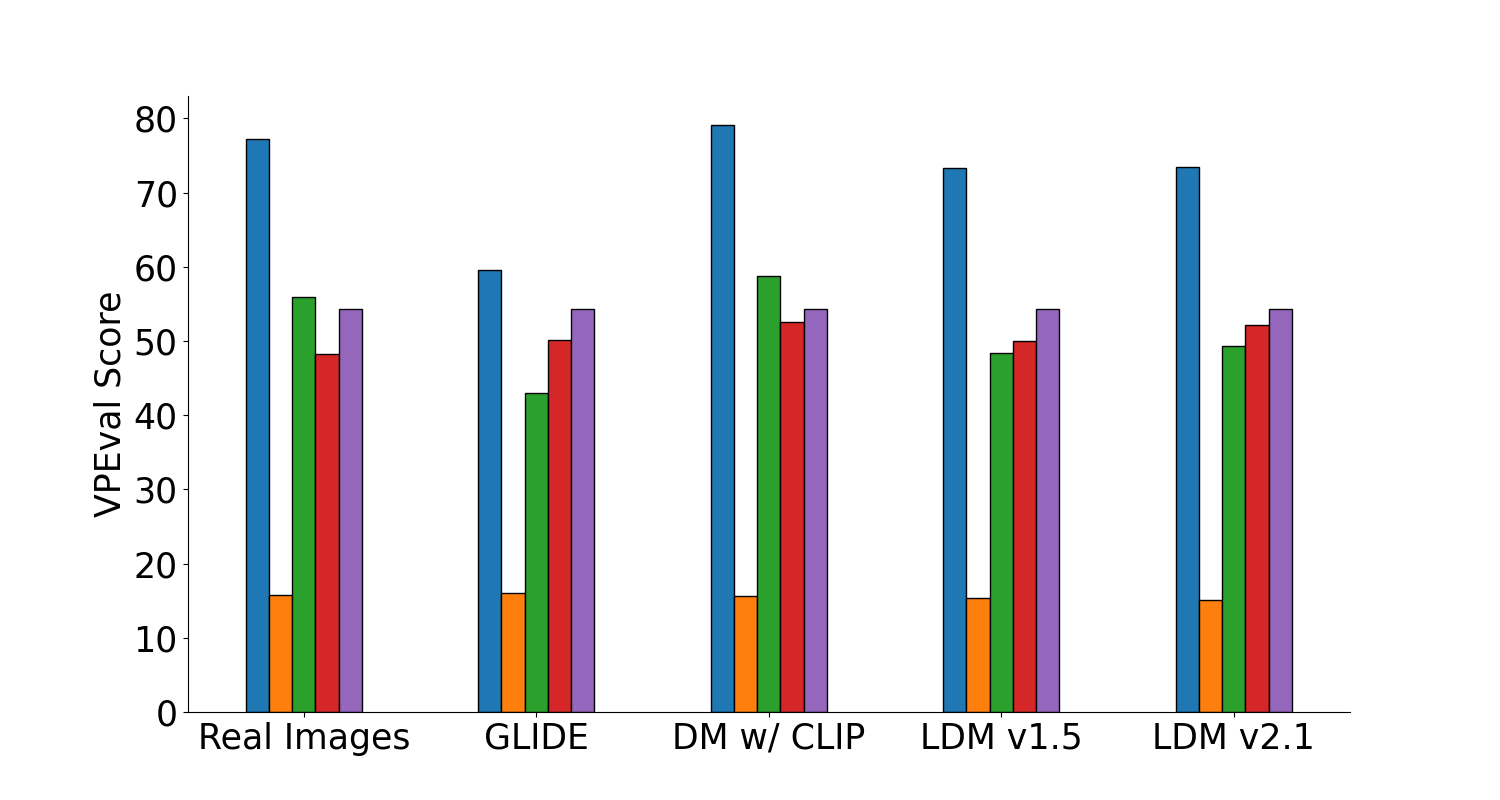}
    \end{subfigure}
    \hspace*{0em}
    \begin{subfigure}{0.46\textwidth}
        \includegraphics[width=1.1\textwidth]{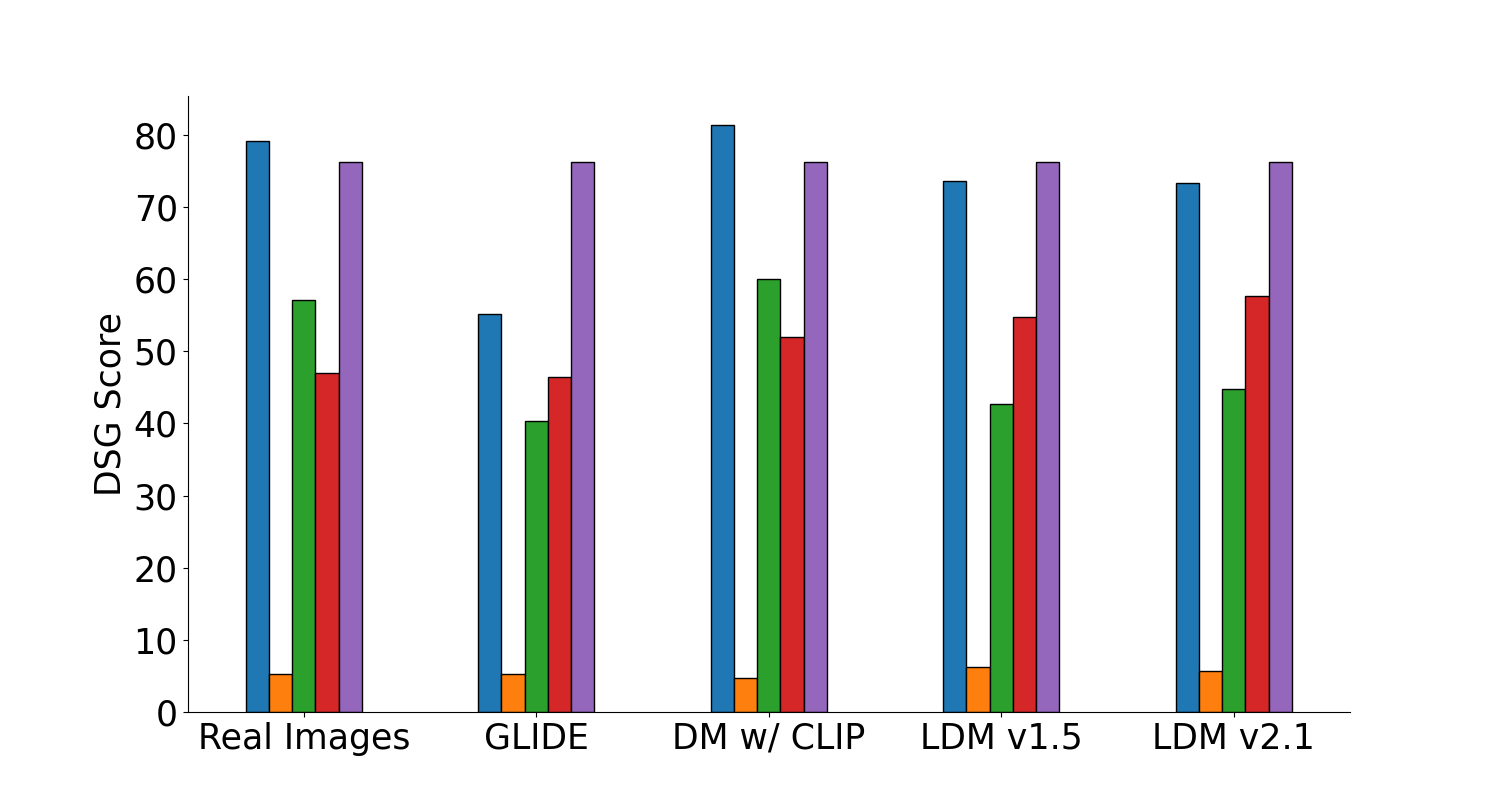}
    \end{subfigure}
    
    \caption{
    Ablation results for COCO.
    The bars refer to the original, unmodified
    metrics {\barplotBlue\bf in blue}, shuffled images (\texttt{Ablation \#1}) {\barplotOrange\bf in orange}, shuffled text (\texttt{Ablation \#2}) {\barplotGreen\bf in green}, using CLIP in place of the VQA model (\texttt{Ablation \#3}) {\barplotRed\bf in red} and using text-only question answering {\barplotPurple\bf in purple} (\texttt{Ablation \#4}). For all metrics, higher is better. While all of the metrics are robust to using random images and to shuffling text, there is a much smaller gap when we ablate the VQA model and instead use CLIP as a proxy (see Figure \ref{fig:vqa-using-clip-questions} for an example).
    }
    \label{fig:ablations-barplot-coco}
\end{figure*}

\section{Ablations: Filling in the Gaps}\label{sec:ablations}

As we mentioned above, a strong evaluation for T2I models should be sensitive to both the information in the text prompt and the corresponding information in the generated image, but our experiments so far suggest that existing metrics are not actually sufficiently influenced by both text and image. For example, in Sections \ref{sec:corr-ling} and \ref{sec:corr-visual}, we found that the metrics have moderate to strong correlations with different linguistic properties of the prompt such as readability and prompt length, and much weaker correlations with visual properties like imageability. To explore this further, we perform four ablation analyses to evaluate the degree to which the visual input is leveraged.

Our ablations target the following hypotheses and predictions:
\begin{itemize}[noitemsep,nolistsep]
    \item[--] \texttt{Ablation 1:} For each example, we select a random image. If there is no performance degradation, we can conclude that  the metrics rely only on the text.
    \item[--] \texttt{Ablation 2:} We reorder the text in the caption (CLIPScore) or questions (TIFA, VPEval, DSG). If there is no performance degradation, we can conclude the metric operates like a bag-of-words.
    \item[--] \texttt{Ablation 3:} We determine whether the VQA model is genuinely necessary by running a pseudo-VQA using CLIP instead. If there is no performance degradation, we can conclude that using a cheaper CLIP alternative is fine. % generating embeddings of the image paired with each of the QA-pairs using CLIP
    \item[--] \texttt{Ablation 4:} Instead of \textit{visual} question answering, we run \textit{text-only} question-answering model using a SOTA QA model. If there is little or no performance degradation, we can conclude that the image is not very important for the metric.
\end{itemize}

Below, we describe the results of our ablations in turn (also see \autoref{fig:ablations-barplot-coco}):

% We find that, while TIFA, VPEval, and DSG do rely on the images to some extent (see \TODO{}), none of these metrics reliably exceed basic yes-only and choice A-only baselines. 

\paragraph{Ablation 1: Shuffled Images.}

We completely ablate the relationship between the images and 
text by randomly selecting an image from the dataset for each example.
Because this shuffling should generally ablate any relationship between questions and images, we expect \metrics to be at or below random chance performance. We observe a huge drop in performance for every metric. This implies that the VQA models generating the \metric scores do not completely ignore the visual input. When an image is completely irrelevant, it can throw the VQA model off and solicit a ``no'', but that doesn't necessarily mean the VQA is sufficiently attending to the image (see Ablation \#4 below). 

\paragraph{Ablation 2: Shuffled Text.}

Instead of shuffling the text \textit{between} examples, we instead shuffle the text \textit{within} an example \citep{gauthier-levy-2019-linking,sinha2021masked}. The order of the words in each question changes, such that a question ``\texttt{What are the animals in the image?}'' may become ``\texttt{image animals are what in the the?}''. Ideally, a strong metric should be sensitive to word order, % as well as the presence of words,
since it can matter (\eg ``\texttt{is the dog to the left of the cat?}'', \citealt{thrush2022winoground}). All four metrics perform worse on this ablation, meaning they are somewhat sensitive to word order, with CLIPScore being the least sensitive, acting mostly as a bag-of-words.

\paragraph{Ablation 3: Running VQA using CLIP.}%\label{sec:ablation-clip-as-vqa}

For the VQA-based metrics, we replace the VQA model with CLIP to understand how much VQA itself contributes to the pipeline. For each question with \textit{N} potential answer choices, we create \textit{N} captions formatted as ``\texttt{\{question\}? \{answer choice n\}}". % Each caption therefore represents the question paired with each possible answer choice.
We use CLIP to compute the cosine similarity between the generated image and each of the \textit{N} captions; see example in Figure \ref{fig:vqa-using-clip-questions}. We treat the CLIP prediction as correct if the caption with the highest cosine similarity is the caption containing the correct answer. 
Performance does degrade for the metrics for all T2I models tested, approximately as much as for the shuffled text ablation, suggesting that QA is necessary.

\paragraph{Ablation 4: Running VQA without the V -- Text-only Question Answering.}

We ablate the visual input entirely by using a text-only LM for QA. Prior work has shown that VQA models heavily rely on textual priors and can even ignore visual input 
\citep{jabri2016revisiting,goyal2017making,agrawal2018CVPR}.
%Due to the unbalanced question distributions,
We explore whether these \metrics may also ignore the images.
%VQA models are used because they explicitly take image input, but to 
To do so, we replace the VQA models with an LM prompted for QA, specifically Flan-T5-XL. We use the same input from the VQA model, formatted as: ``\texttt{Question: \{question\} Choices: \{choices\} Answer:}". We find that text-only QA performed fairly well, just shy of metrics using VQA. This suggests that VQA may not be strictly necessary: Because the generated questions are likely very skewed towards very probable answers, text-only QA appears to be basically sufficient.

\section{Discussion and Conclusion}\label{sec:discussion}

\paragraph{Other desiderata for automatic \metricsNoSpace.} Ideally, a metric would satisfy all minimal necessary desiderata, and several nice-to-haves as well. There are many additional nice-to-haves that also exist, such as sufficient generation diversity \citep{hall2023dig}, robustness to minor input perturbations \citep{jiang-etal-2020-know, gao-etal-2021-making, sinha-etal-2021-perturbing, goodarzi-etal-2023-robustness}, sensitivity to input sample difficulty, etc. Probably most relevant to this work on automatic \metrics is compute efficiency. While chaining together submodules -- such as LMs, VLMs, or VQA systems -- is promising \citep{manas2024improving}, incorporating these subcomponents into model pipelines can add additional compute costs. High compute costs during training have been tied to environmental consequences \citep{strubell-etal-2019-energy}, and it is also possible that incorporating submodules such as LMs into our evaluation pipelines during inference may also have such consequences. Concurrent to our work, \citet{saxon2024evaluates} performed a different type of meta-evaluation and also showed that VQA-based metrics that use additional submodules may not outperform simpler embedding space metrics like CLIPScore. This is why it very important to demonstrate additional utility when proposing novel metrics, especially when they rely on expensive subcomponents.

\paragraph{Conclusion.} We defined a set of desiderata that should be considered when designing new \metrics for text-to-image models. We analyzed 4 metrics -- CLIPScore, TIFA, VPEval and DSG -- and found they  struggle to meet all desiderata. First, instead of using both the textual and visual information, they  rely much more on the text. Next, excluding CLIPScore, they have a very strong correlation with one another, calling into questions how much new information is contributed by each successive metric. Lastly, the VQA-based metrics (TIFA, VPEval and DSG) have very skewed question distributions with artifacts that makes it difficult to know whether they are genuinely measuring text-image consistency. We hope our desiderata can be useful in ensuring we are designing robust evaluation metrics as the field of text-to-image generation continues to grow.

% \section{Acknowledgments} Michal Drozdzal, Hagen Blix, Pietro? Oscar?

\bibliography{colm2024_conference, anthology}
\bibliographystyle{colm2024_conference}

\appendix
\section{\Metric Correlations for Winoground}

\subsection{Correlations between Linguistic and Visual Properties}

We present the results for the correlation between different linguistic and visual properties for \metrics in Table \ref{tab:ling-corr-winoground} and \ref{tab:visual-corr-winoground}. Similar to the findings for COCO, we observe moderate to strong correlation for some of linguistic properties and essentially no correlation for the visual properties. Unlike COCO, we do not observe correlations for readability and complexity. We hypothesize that this may be because some of the captions in Winoground are less typical (\eg the short captions \textit{truck fire} and \textit{fire truck}).

\renewcommand{\tableCellLing}[1]{\rotatebox{80}{\parbox{4em}{ #1}}}
\renewcommand{\tableCellVisual}[1]{\rotatebox{80}{\parbox{4em}{ #1}}}

\begin{table*}[!ht]
\centering

\begin{subtable}{\textwidth}
\centering
% \textbf{\small Spearman's Rank Correlation $\rho$ between Metrics and Linguistic Properties} \\
    
\scalebox{0.8}{
\begin{tabular}{l|llll |llll| llll}
    & \multicolumn{4}{c}{$\rho$ -- Readability}
    & \multicolumn{4}{c}{$\rho$ -- Complexity}
    & \multicolumn{4}{c}{$\rho$ -- Length} \\

    & \tableCellLing{CLIPScore}
    & \tableCellLing{TIFA}
    & \tableCellLing{VPEval}
    & \tableCellLing{DSG}

    & \tableCellLing{CLIPScore}
    & \tableCellLing{TIFA}
    & \tableCellLing{VPEval}
    & \tableCellLing{DSG}
    
    & \tableCellLing{CLIPScore}
    & \tableCellLing{TIFA}
    & \tableCellLing{VPEval}
    & \tableCellLing{DSG} \\
    
    \toprule
    Real Images & 0.06 & -0.05 & 0.10 & -0.10 & 0.06 & -0.10 & 0.02 & -0.18 & 0.06 & -0.26 & -0.05 & \bf -0.76* \\
    
    \glide & 0.04 & 0.08 & -0.04 & -0.10 & 0.05 & 0.05 & 0.02 & -0.19 & 0.29 & 0.06 & 0.03 & -0.36 \\

    \dmwithclipshort & 0.14 & -0.04 & 0.32* & 0.11 & 0.16 & 0.00 & 0.02 & -0.10 & \bf 0.67* & \bf -0.38 & \bf 0.44 & \bf -0.70* \\

    \ldmOne & -0.12 & -0.14 & -0.27 & -0.28 & 0.16 & -0.14 & -0.13 & -0.29* & 0.06 & \bf -0.51* & -0.32 & \bf -0.80* \\
        
    \ldmTwo & 0.04 & 0.37* & 0.16 & 0.37* & -0.08 & -0.20 & -0.20 & -0.23 & \bf 0.45 & 0.08 & \bf 0.50* & -0.36 \\
    \bottomrule
\end{tabular}
}
\caption{Spearman's rank correlation between \textit{linguistic features} and \metrics on the Winoground dataset.}
\label{tab:ling-corr-winoground}
\end{subtable}

\begin{subtable}{\textwidth}
\centering
% \textbf{\small Spearman's Rank Correlation $\rho$ between Metrics and Visual Properties} \\

\scalebox{0.85}{
\begin{tabular}{l| llll| llll| llll}

    \multicolumn{1}{c}{}
    & \multicolumn{4}{c}{$\rho$ -- Concreteness}
    & \multicolumn{4}{c}{$\rho$ -- Imageability}
    & \multicolumn{4}{c}{$\rho$ -- IN21k Caption Overlap}\\

    & \tableCellVisual{CLIP}
    & \tableCellVisual{TIFA}
    & \tableCellVisual{VPEval}
    & \tableCellVisual{DSG}
    
    & \tableCellVisual{CLIP}
    & \tableCellVisual{TIFA}
    & \tableCellVisual{VPEval}
    & \tableCellVisual{DSG}
    
    & \tableCellVisual{CLIP}
    & \tableCellVisual{TIFA}
    & \tableCellVisual{VPEval}
    & \tableCellVisual{DSG} \\
    
    \toprule
    
    Real Images & 0.06 & -0.05 & -0.02 & -0.03 & 0.15* & 0.00 & 0.04 & 0.01 & 0.24 & -0.14 & -0.25 & -0.15 \\
    
    \glide & 0.08 & -0.05 & -0.06 & -0.01 & 0.13 & -0.11 & -0.00 & 0.01 & 0.02 & -0.51* & -0.30 & -0.08 \\
    
    %$\backslash$glidebow & 0.01 & -0.08 & -0.04 & -0.05 & 0.12 & -0.04 & 0.02 & 0.01 & 0.01 & -0.55* & -0.19 \\ \hline
    
    \dmwithclipshort & -0.01 & 0.01 & 0.03 & 0.02 & -0.01 & 0.04 & 0.02 & 0.06 & 0.12 & -0.13 & -0.02 & 0.16 \\
    
    \ldmOne & 0.06 & 0.01 & 0.05 & 0.08 & 0.25* & 0.09 & 0.15* & 0.17* & 0.35 & -0.35 & -0.16 & 0.01 \\
    
    \ldmTwo & -0.01 & 0.05 & 0.04 & 0.09 & 0.09 & -0.00 & 0.05 & 0.05 & 0.24 & -0.32 & -0.23 & -0.01 \\
    \bottomrule
    \end{tabular}
}
\caption{Spearman's rank correlation between \textit{visual features} and \metrics on the Winoground dataset.}
\label{tab:visual-corr-winoground}
\end{subtable}

\caption{Spearman's Rank Correlation between \metrics and different linguistic properties (top) and visual properties (bottom) for the Winoground dataset. Winoground shows generally simialr trend to COCO, with smaller magnitudes; see Table \ref{tab:corr-coco}. These findings also support that the metrics are more language-related than vision related.}
\label{tab:corr-winoground}
\end{table*}

\subsection{Correlation between Metrics}
\begin{figure*}
    \centering
    
    \begin{tikzpicture}
    % top row
    \node(real-images){
        \includegraphics[width=17em]{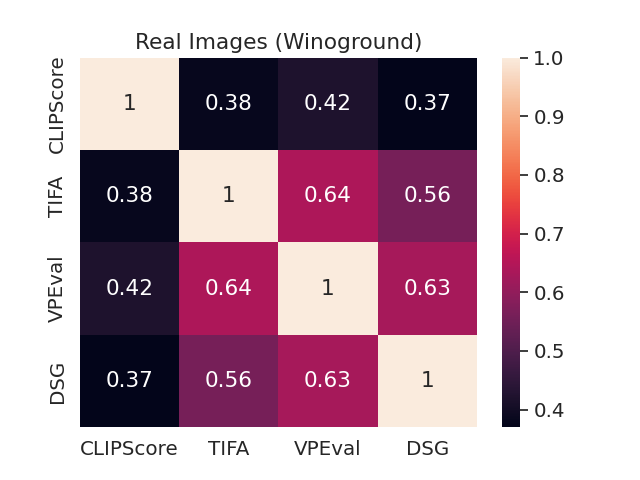}
    };
    \node[right=-5em of real-images](dalle2){
        \includegraphics[width=17em]{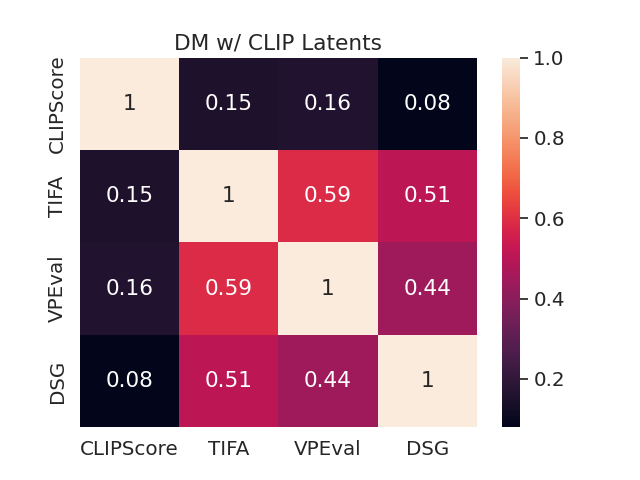}
    };
    \node[right=-5em of dalle2](glide){
        \includegraphics[width=17em]{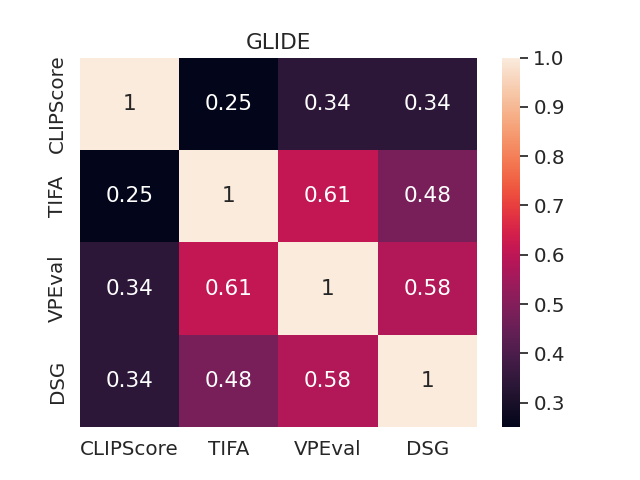}
    };
    % bottom row
    %\node[below=-1ex of real-images](glide-bow){
    %    \includegraphics[width=17em]{winoground_heatmap_glide_bow.png}
    %};
    \node[below=-1ex of dalle2,xshift=-7em](sd-hf){
        \includegraphics[width=17em]{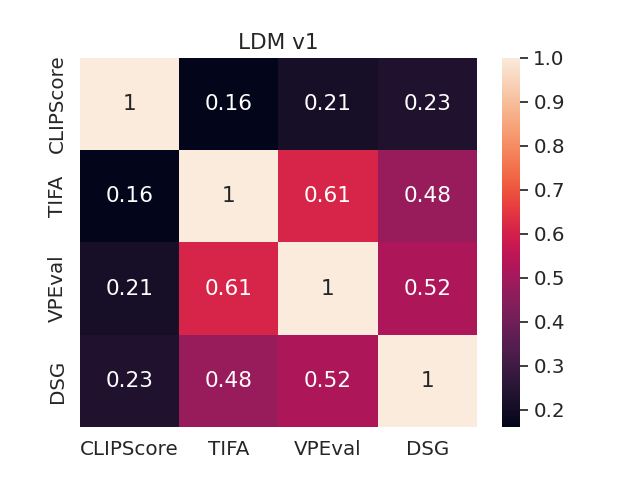}
    };
    \node[below=-1ex of glide,xshift=-7em](sd-api){
        \includegraphics[width=17em]{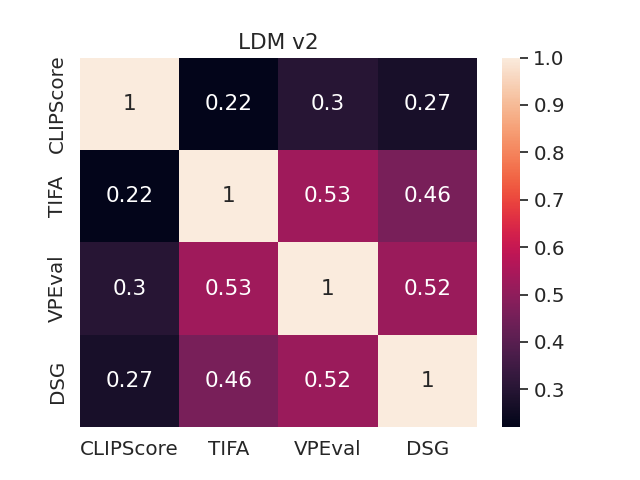}
    };
    \end{tikzpicture}
    
    \caption{Correlation between each pair of \metrics -- CLIPScore, TIFA, VPEval and DSG -- for 4 text-to-image generative models and for real images. Similar to the finding for COCO shown in Figure \ref{fig:exp3-coco}, we find that VQA-based metrics do not correlate with CLIPScore (again with the exception of \glideNoSpace). For VQA-based metrics on the other hand, correlations are medium to strong and statistically significant. This suggests that the contributions from each metric may not be that distinct from the other metrics. Again similar to COCO, we also observe similar patterns between the real images (top left) and the generated images, suggesting the metrics are likely to be consistent across different image sources such as new text-to-image models. See Section \ref{sec:metrics-distinct} for more details.}
    \label{fig:exp3-winoground}
\end{figure*}

Next, in Figure \ref{fig:exp3-winoground}, we present the correlation between metrics as described in Section \ref{sec:metrics-distinct}. Similar to COCO, we find moderate to strong correlations between the VQA-based \metrics. We also find that these VQA-based metrics do not have a strong correlation with CLIPScore.

\section{Ablation Results on Winoground}
\begin{figure*}[h!]
    \centering
    \begin{subfigure}{0.9\textwidth}
        \includegraphics[width=\textwidth]{ablation_legend_one_row.png}
    \end{subfigure}

    \hspace*{-3em}
    \begin{subfigure}{0.48\textwidth}
        \includegraphics[width=1.2\textwidth]{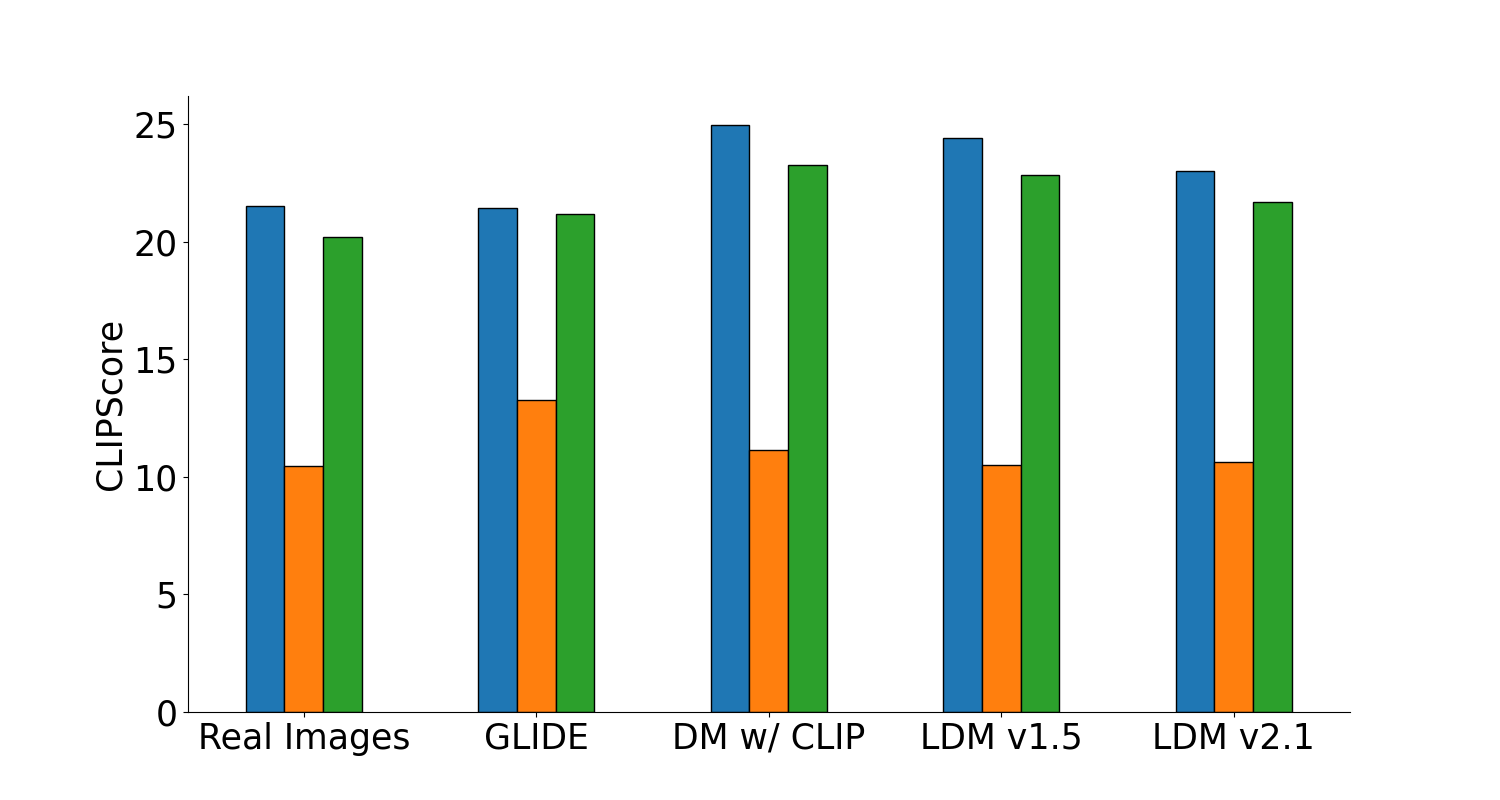}
    \end{subfigure}
    \hspace*{-2em}
    \begin{subfigure}{0.48\textwidth}
        \includegraphics[width=1.2\textwidth]{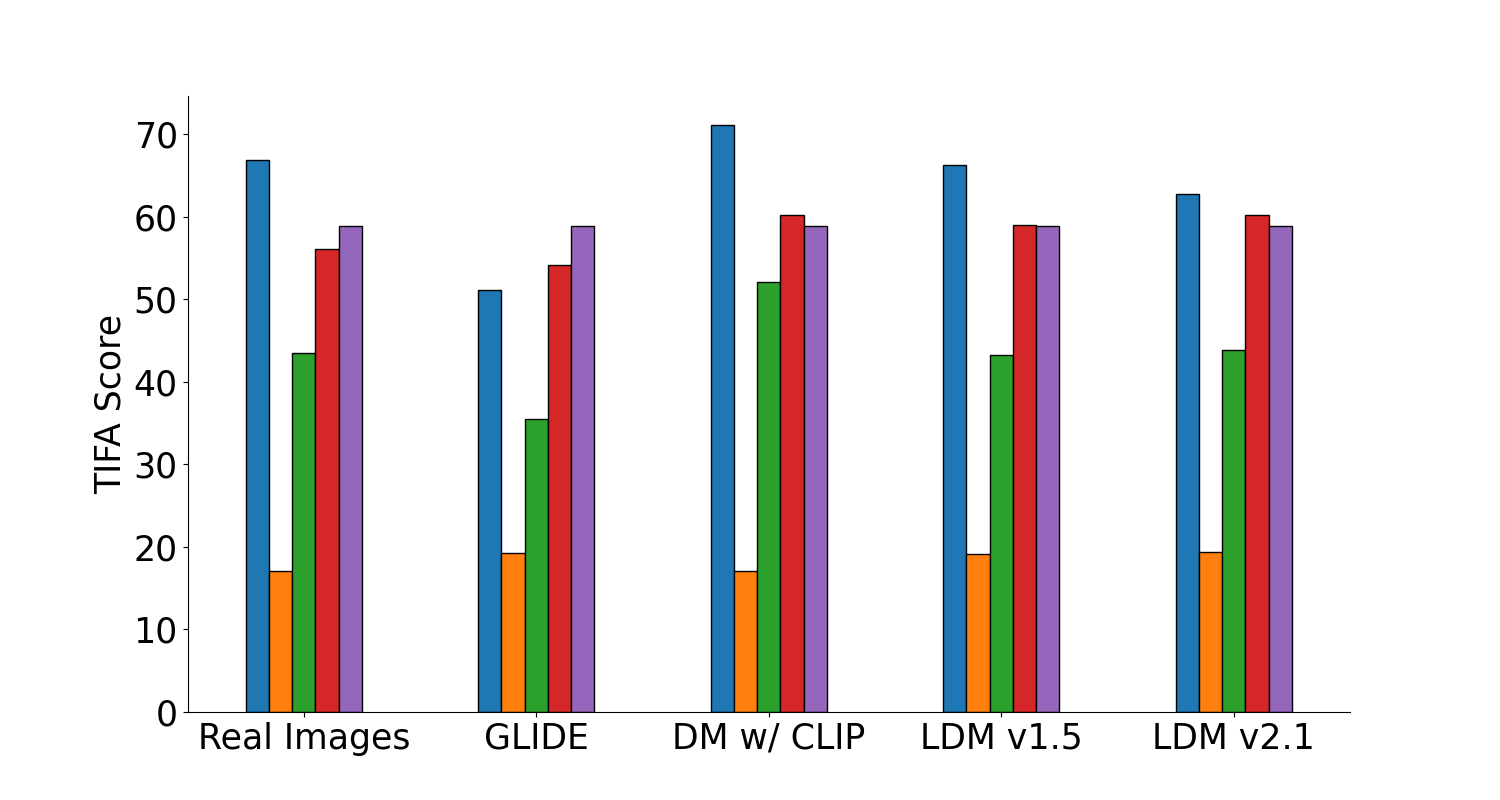}
    \end{subfigure}
    \\

    \hspace*{-3em}
    \begin{subfigure}{0.48\textwidth}
        \includegraphics[width=1.2\textwidth]{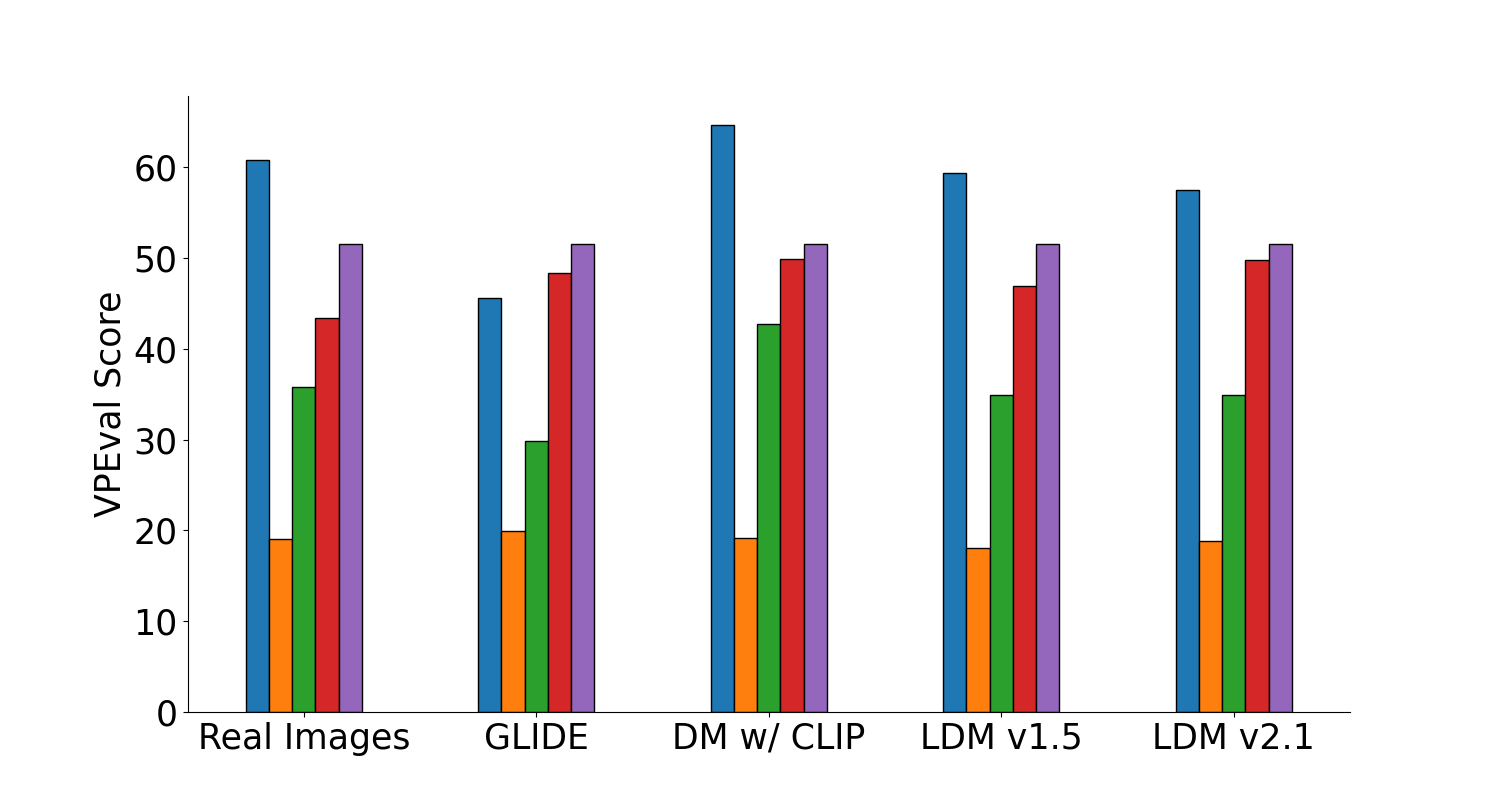}
    \end{subfigure}
    \hspace*{-2em}
    \begin{subfigure}{0.48\textwidth}
        \includegraphics[width=1.2\textwidth]{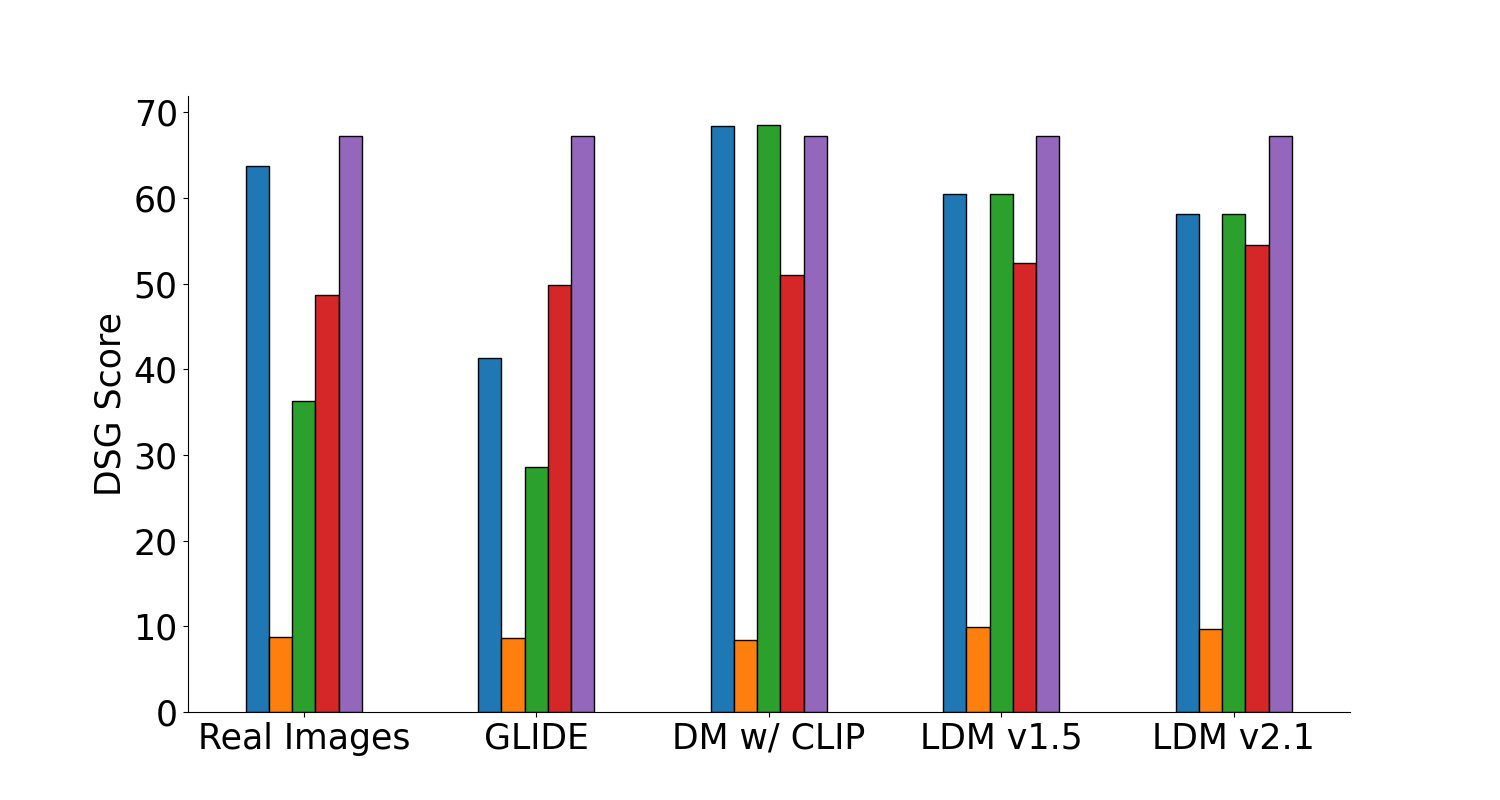}
    \end{subfigure}
    
    \caption{Ablation results for Winoground dataset.
    The bars refer to the original, unmodified
    \metrics {\barplotBlue\bf in blue}, shuffled images (\texttt{Ablation \#1}) {\barplotOrange\bf in orange}, shuffled text (\texttt{Ablation \#2}) {\barplotGreen\bf in green}, using CLIP in place of the VQA model (\texttt{Ablation \#3}) {\barplotRed\bf in red} and using text-only question answering {\barplotPurple\bf in purple} (\texttt{Ablation \#4}). For all metrics, higher is better. % While all of the metrics are robust to using randomly ordered text as questions and to using random images, there is a much smaller gap when we ablate the VQA model and instead use CLIP as a proxy (see Figure \ref{fig:vqa-using-clip-questions}
    }
    \label{fig:ablations-barplot-winoground}
\end{figure*}

\begin{figure}[h!]
    \centering
    \centering
    % \resizebox{1\textwidth}{!}{%
    \begin{tikzpicture}
        % \tikzstyle{every node}=[font=\small]
        \node(ques) {
            \scalebox{0.7}{
            \parbox{30em}{\texttt{
            \hspace{-0.5em}{\bf Question:} What type of animal is this animal?\\
            {\bf Choices:} \{dog, cat, bird, fish\}\\
            {\bf Answer:} dog\\[3ex]
            {\bf Captions for CLIP:}\\
            c$_{\textrm{\texttt{0}}}$ = What type of animal is this animal? dog\\
            c$_{\textrm{\texttt{1}}}$ = What type of animal is this animal? cat\\
            c$_{\textrm{\texttt{2}}}$ = What type of animal is this animal? bird\\
            c$_{\textrm{\texttt{3}}}$ = What type of animal is this animal? fish\\[2ex]
            {\bf Accuracy:} Let s(c$_{\textrm{\texttt{i}}}$) be the CLIPScore for a caption c$_{\textrm{\texttt{i}}}$. If the caption containing the correct choice, s(c$_{\textrm{\texttt{0}}}$), is the highest score among the captions, then this question is correct.
            }}}
        };
        \node[left=2em of ques]{
        \includegraphics[width=12em]{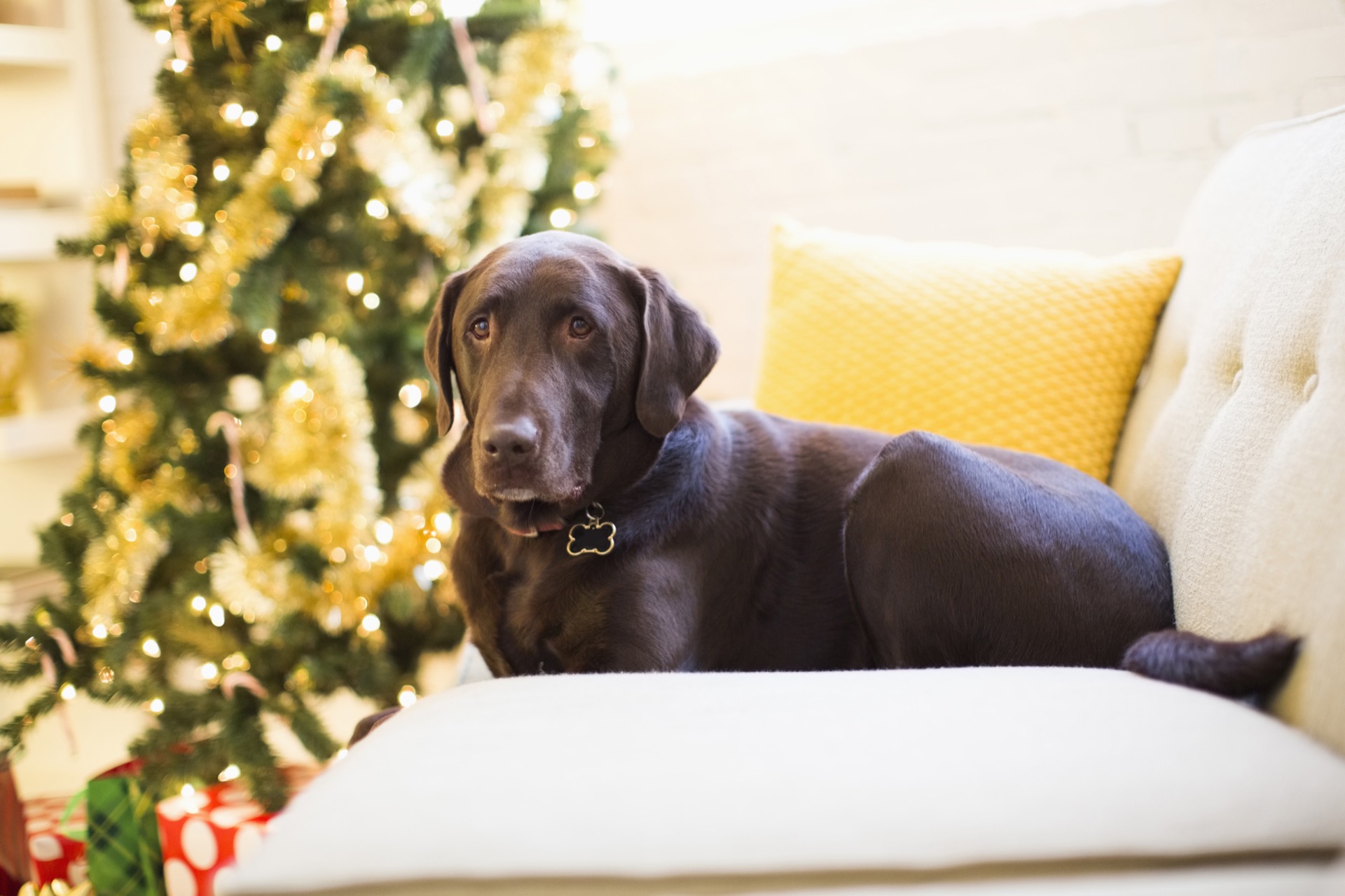}
        };
    \end{tikzpicture}
    \vspace*{-2ex}
    
    \caption{For the TIFA, VPEval and DSG \metricsNoSpace, we use ablate the VQA model and replace it with CLIP. Every question is paired with a set of choices and a ground-truth answer \textbf{(top)}. The question and choices are combined to form captions, where \textit{N} choices yields \textit{N} captions \textbf{(middle)}. Finally, the CLIPScore is computed for each of the \textit{N} captions. If the CLIPScore for the ground-truth caption is the highest among the \textit{N} captions, then the question is marked as correct \textbf{(bottom)}. %; see Section \ref{sec:ablation-clip-as-vqa} for more details.
    }
    \label{fig:vqa-using-clip-questions}
\end{figure}

We show the results of our four ablations on Winoground in Figure \ref{fig:ablations-barplot-winoground}. The ablations including shuffling the images between examples (\texttt{Ablation \#1}), shuffling the text within a given question/caption (\texttt{Ablation \#2}), using CLIP in place of the VQA model (\texttt{Ablation \#3}) and lastly using a text-only model for question answering in place of the VQA model (\texttt{Ablation \#4}). An example of the format for \texttt{Ablation \#3} where we use CLIP for VQA is shown in Figure \ref{fig:vqa-using-clip-questions}.

\section{Statistics on Question Distribution}
In Section \ref{sec:textartifact}, we dug deeper into the generated questions for TIFA, VPEval and DSG. CLIPScore does not use generated questions, instead using the caption directly, and therefore is not included. Table \ref{tab:qas-stats} shows the more detailed statistics of the question distributions. We observe that the distribution is quite skewed with nearly every \textit{yes-no} question having a ground-truth answer of \textit{yes}. Additionally, nearly every multiple-choice question has a first answer bias.

We know the \textit{yes}-bias and \textit{first-correct} bias are a spurious correlations that impacts LMs and/or (V)QA models. Moreover, these spurious lexical correlations from the LM generating the questions and the VQA model answering these questions could compound. Say the LM often writes questions containing the word ``bear'' and answers them with ``yes'', regardless of whether the prompt contains ``bear'' or the generated image contains a bear. Also imagine the VQA model often says ``yes'' to questions containing ``bear''. In this case, no matter what the input is, the LM will talk about bears and expect a ``yes'', and the VQA model will provide it. Surely, this is an extreme toy example, but past work suggests LMs \citep{shwartz-etal-2020-grounded, tu-etal-2020-empirical, smith2021hi, goodarzi-etal-2023-robustness} and VQA models \citep{agrawal2018CVPR, ray2019sunny, shah2019cycle, agarwal2020towards,sheng2021human} still suffer from artifacts. If we want to chain models together, we need to think hard about which kinds of spurious correlations may exist.

\renewcommand{\leftmostCell}[1]{\scalebox{1}{\parbox{15em}{#1}}}
\begin{table*}
    \centering\hspace*{-1em}
    \scalebox{0.98}{
    \begin{tabular}{l | r|r|r ||r|r|r}
    \multicolumn{1}{c|}{}
    & \multicolumn{3}{c}{\bf COCO} &
    \multicolumn{3}{c}{\bf Winoground} \\[1.2ex]
    \toprule
    & \tableCell{TIFA}
    & \tableCell{VPEval}
    & \tableCell{DSG}
    
    & \tableCell{TIFA}
    & \tableCell{VPEval}
    & \tableCell{DSG}
    % & \tableCell{\small VPEval \\ -- Llama 2\\[-0.5ex]}
    \\\hline
    
    \leftmostCell{Total \# of questions} & 61.2k & 118.4k & 63.6k & 1.8k & 7.3k & 1.3k\\

    \leftmostCell{What \% are yes/no questions? ($\%$)\\[-1.2ex]}
    & 56.8 & 61.5 & 100 & 61.0 & 63.6 & 100 \\

    \leftmostCell{What \% is correct answer \textit{yes}?\\[-1.2ex]}
    & 99.7 & 99.2 & 100 & 99.3 & 96.6 & 100 \\

    % \parbox{12em}{$\rightarrow$ Of yes/no Qs, how often is the correct answer \textit{no}? ($\%$)\\[-1.2ex]}
    % &  & 0.27 & 0.82 & 1.68 & 3.53 \\

    \leftmostCell{What \% is correct answer \textit{no}?\\[-1.2ex]}
     & $<$1 & $<$ 1 & 0 & 1.7 & 3.5 & 0 \\

    \leftmostCell{What \% are multiple choice?\\[-1.2ex]}
    & 43.2 & 38.5 & 0 & 39.0 & 34.4 & 0 \\

    \leftmostCell{What \% is correct answer the 1st choice?\\[-1.2ex]}
    & 94.0 & 93.8 & N/A & 92.0 & 89.7 &  N/A\\
    \bottomrule
    \end{tabular}
    }
    \caption{Statistics on the questions generated by an LM for VQA portion of TIFA, VPEval and DSG. For yes/no questions, the correct answer is \textit{almost always yes} ($\thicksim$99\% of the time). For multiple choice questions (excludes DSG because all questions are binary yes/no), \textit{the first answer is almost always correct}. Overall, the distribution of LM-generated questions for all \metrics are highly skewed.}
    \label{tab:qas-stats}
\end{table*}

\end{document}